\documentclass[11pt]{article}

\usepackage[final]{acl}
\usepackage{multirow}
\usepackage{float}
\usepackage{times}
\usepackage{latexsym}
\usepackage{amsmath}
\usepackage[most]{tcolorbox}
\usepackage[T1]{fontenc}

\usepackage[utf8]{inputenc}
\usepackage{amssymb}
\usepackage{newunicodechar}
\newunicodechar{：}{:}
\usepackage{algorithm}
\usepackage{algpseudocode}
\usepackage{wrapfig}

\usepackage{microtype}

\usepackage{inconsolata}

\usepackage{graphicx}
\usepackage{subcaption} 
\usepackage{svg}      

\usepackage{array}
\usepackage{xcolor}
\usepackage{colortbl}
\usepackage{longtable}
\usepackage{booktabs}
\usepackage{arydshln}
\usepackage{hyperref}

\usepackage{appendix}
\newcommand{\ourMethod}{TRiMS\space}

\title{TRiMS: Real-Time Tracking of Minimal Sufficient Length for Efficient Reasoning via RL}

\author{
  \textbf{Tingcheng Bian\textsuperscript{1,2}}, \textbf{Jinchang Luo\textsuperscript{1}}, \textbf{Mingquan Cheng\textsuperscript{1}}, \textbf{Jinyu Zhang\textsuperscript{3}},
  \\
  \textbf{Xiaoling Xia\textsuperscript{1}}, \textbf{Ni Li\textsuperscript{1}}, \textbf{Yan Tao\textsuperscript{1}}, \textbf{Haiwei Wang \textsuperscript{1}}\thanks{Corresponding authors. Email: \href{mailto:wanghaiwei@baidu.com}{wanghaiwei@baidu.com}}
\\
  \textsuperscript{1}Baidu Inc.
  \\
  \textsuperscript{2}Shenzhen University
  \\
   \textsuperscript{3}Harbin Institute of Technology
  \\
  }

\definecolor{questionbg}{RGB}{245,245,245}
\definecolor{correctgreen}{RGB}{200,235,200}
\definecolor{graybg}{RGB}{240,240,240}
\definecolor{greenbg}{RGB}{235,245,235}

\begin{document}
\maketitle

\begin{abstract}
Large language models achieve breakthroughs in complex reasoning via long chain-of-thought sequences. However, this often leads to severe reasoning inflation, causing substantial computational redundancy. To maximize Intelligence per Token, we introduce a theoretical metric, MSL-Minimal Sufficient Length. MSL rigorously characterizes the shortest reasoning length that preserves answer correctness. We provide a recursive definition based on independently sampled sequences and prove the existence of its limit, establishing the first measurable lower bound for reasoning-chain compression. Building on an analysis of mainstream CoT compression strategies, we identify key structural factors enabling a model to approach MSL. Based on these insights, we propose \ourMethod which employs the GRPO algorithm in conjunction with MSL-based estimation during training, while mitigating instabilities during the training process through dynamic batch aggregation and advantage computation using batch-level standard deviation. \ourMethod achieves over 80\% CoT token reduction with a minor accuracy boost across all benchmarks.

\end{abstract}

\section{Introduction}
{

\begin{figure}[!t]
  \centering
  \begin{subfigure}[t]{\columnwidth}
    \centering
    \includegraphics[width=0.9\linewidth]{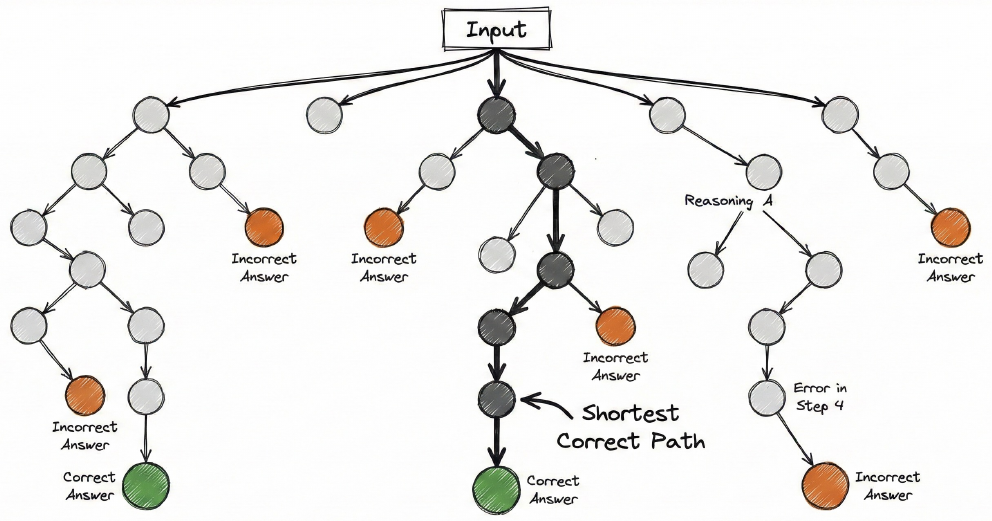}
    \label{fig:intro-msl-top}
  \end{subfigure}

  \vspace{2pt} 

  \begin{subfigure}[t]{\columnwidth}
    \centering
    \includegraphics[width=0.95\linewidth]{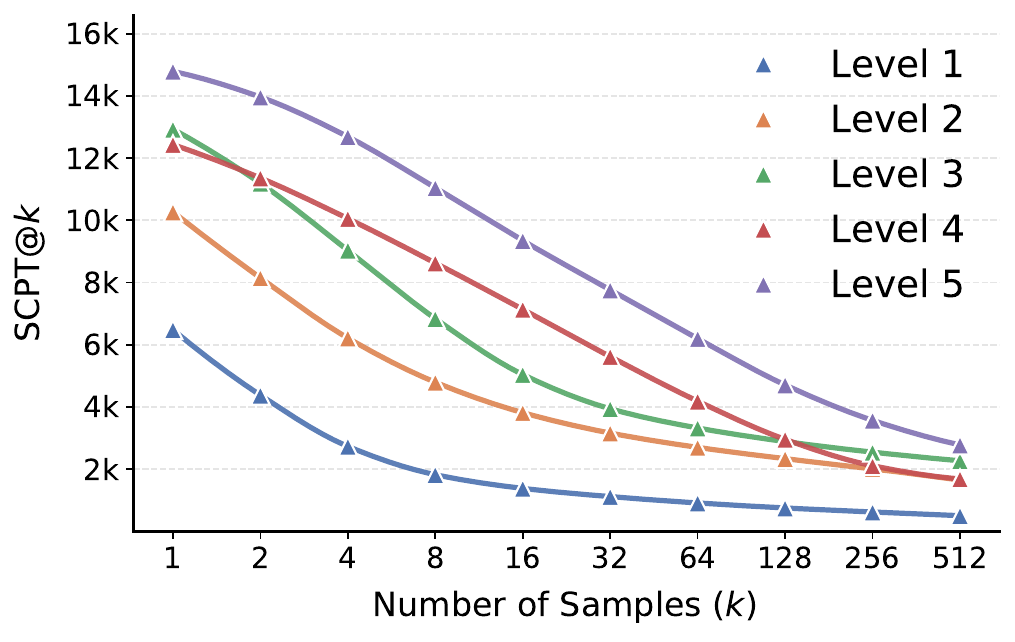}
    \label{fig:intro-msl-bottom}
  \end{subfigure}

  \caption{
\textbf{(Top)} Conceptual illustration of diverse reasoning paths generated by LLM. \textbf{(Bottom)} Expected token length of the shortest correct reasoning path(SCPT@K) as a function of the sampling times $k$. 
  }
  \label{fig:introduction}
\end{figure}

\paragraph{Importance:}

Despite the significant breakthroughs achieved by reasoning-oriented models such as OpenAI-o1, DeepSeek-R1 and Qwen through long CoT reasoning on complex tasks~\cite{jaech2024openai,guo2025deepseek, yang2025qwen3, wei2022chain}, their accompanying high token consumption, inference latency, and severe computational redundancy on simple problems have become critical bottlenecks for large-scale deployment. A central challenge in current research is how to systematically improve intelligence per token~\cite{fu2022complexity}.

Although prior work has attempted to enhance reasoning efficiency via prompt engineering~\cite{renze2024benefits,xu2025chain,ma2025reasoning}, Inference strategy~\cite{xia2502tokenskip,yan2025mur,huang2025efficient}, supervised fine-tuning~\cite{wu2025efficient,zhang2025dart,lu2025retro}, or reinforcement learning~\cite{zhu2025entropy,li2025compressing,wu2026anti}, these approaches often face an inherent trade-off between performance and compression. In the absence of a clear theoretical characterization of the boundary between accuracy and compression ratio~\cite{lee2025well}, they tend to oscillate between over-compression leading to performance collapse~\cite{ma2025cot} and conservative compression with limited gains~\cite{kang2025c3ot}. Fundamentally, the field lacks an optimization target that can directly quantify a model’s intrinsic reasoning capability while remaining strongly coupled with correctness~\cite{wang2023towards}. The core question is therefore: under the constraint of correctness, where does the ultimate compression limit of CoT lie?

\paragraph{Observation:}

To investigate this question, we view autoregressive generation as a multi-branch decision tree (Fig.~\ref{fig:introduction}, top). 
Empirical analysis on the DeepScaleR dataset (see Appendix~\ref{Appendix:A.1} for experimental details) reveals that among all trajectories reaching a correct answer, there exists a naturally emerging shortest path.
As the number of samples $k$ increases, the expected length of this shortest correct path decreases and converges across tasks (Fig.~\ref{fig:introduction}, bottom). This suggests that a model’s reasoning process has an intrinsic limiting lower bound that is approached as the sampling space expands.

\paragraph{Motivation:}

Motivated by these observations, we seek to formalize this limit. Inspired by the statistical formulation of PASS@k~\cite{yue2025does}, we employ combinatorial counting to derive an unbiased estimator for the expected shortest correct length from $k$ independent samples (Appendix~\ref{Appendix:A.2}). Based on this, we define the Minimal Sufficient Length as the limit of tokens required for correct reasoning under a model’s intrinsic distribution. Anchored by MSL, we propose TRiMS, a reinforcement learning framework designed to estimate and approach this theoretical boundary.

\paragraph{Contributions:}
This paper makes the following contributions:

\begin{itemize}  
  \item \textbf{Minimal Sufficient Length.}  
    We introduce MSL, a principled objective that formalizes the shortest reasoning path required for correctness, providing a measurable lower bound for CoT redundancy.

  \item \textbf{TRiMS framework.}  
    We propose \textbf{TRiMS}, an RL-based adaptive compression framework that estimates and approaches MSL online, actively steering the model toward minimal sufficient reasoning.

  \item \textbf{Empirical Validation.}  
    Across multiple benchmarks, TRiMS reduces inference token usage by over $80\%$ while maintaining or improving the accuracy of standard CoT baselines.
\end{itemize}

}

\section{Related Work}


\subsection{Inference-time Strategies}
Recent studies explore dynamic reasoning modulation via mechanisms such as lookahead switching~\cite{zeyu2025large, zhang2025alphaone}, uncertainty-based early exits~\cite{yang2025dynamic, huang2025efficient}, and adaptive routing~\cite{lee2025confidence, huang2025adactrl, wu2025concise}. While effective in reducing average compute, these strategies often incur runtime overhead from auxiliary modules or suffer from unstable truncation criteria~\cite{yang2025specexit}. This limitation highlights the necessity of learning-based compression, which internalizes efficiency directly into the model parameters~\cite{ma2025cot, liu2025conpress, testolina2021learningbased}.

\begin{figure*}[!t]
    \centering

    \begin{minipage}[c]{0.48\textwidth}
        \centering
        
        \begin{subfigure}[t]{0.48\linewidth}
            \centering
            \includegraphics[width=\linewidth]{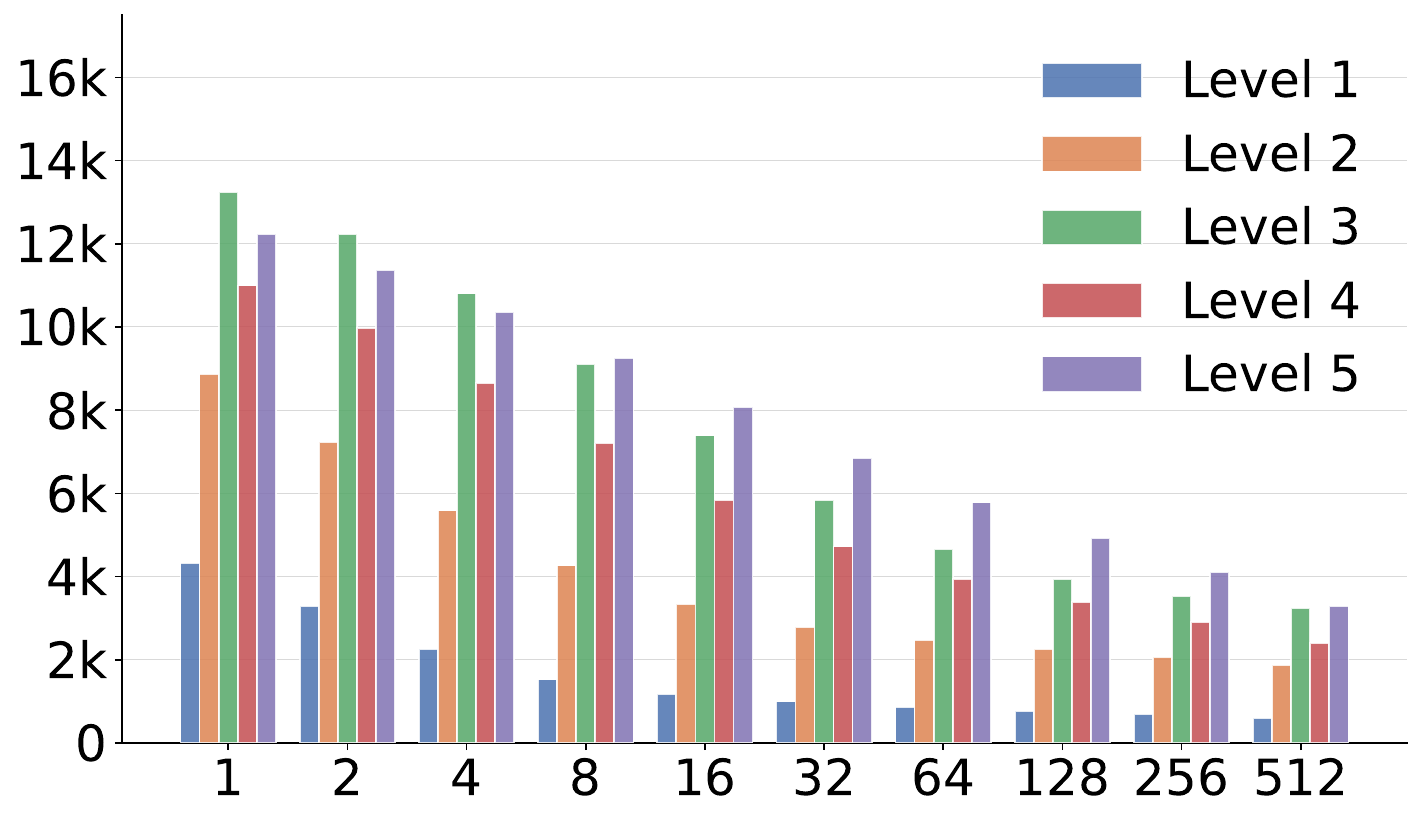}
            \caption{Low-temperature}
        \end{subfigure}
        \hfill
        \begin{subfigure}[t]{0.48\linewidth}
            \centering
            \includegraphics[width=\linewidth]{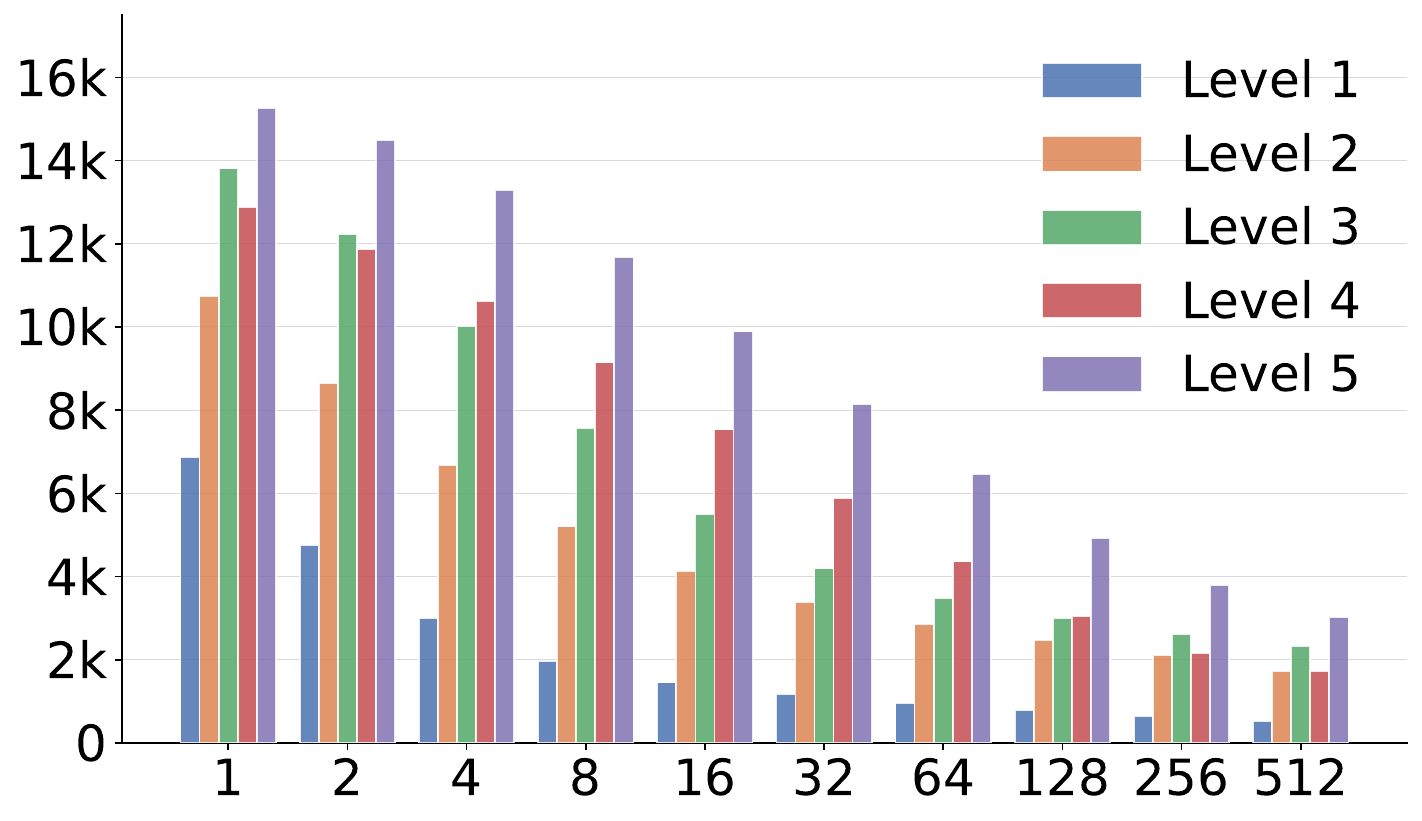}
            \caption{Moderate-temperature}
        \end{subfigure}

        \vspace{0.6em}

        \begin{subfigure}[t]{0.48\linewidth}
            \centering
            \includegraphics[width=\linewidth]{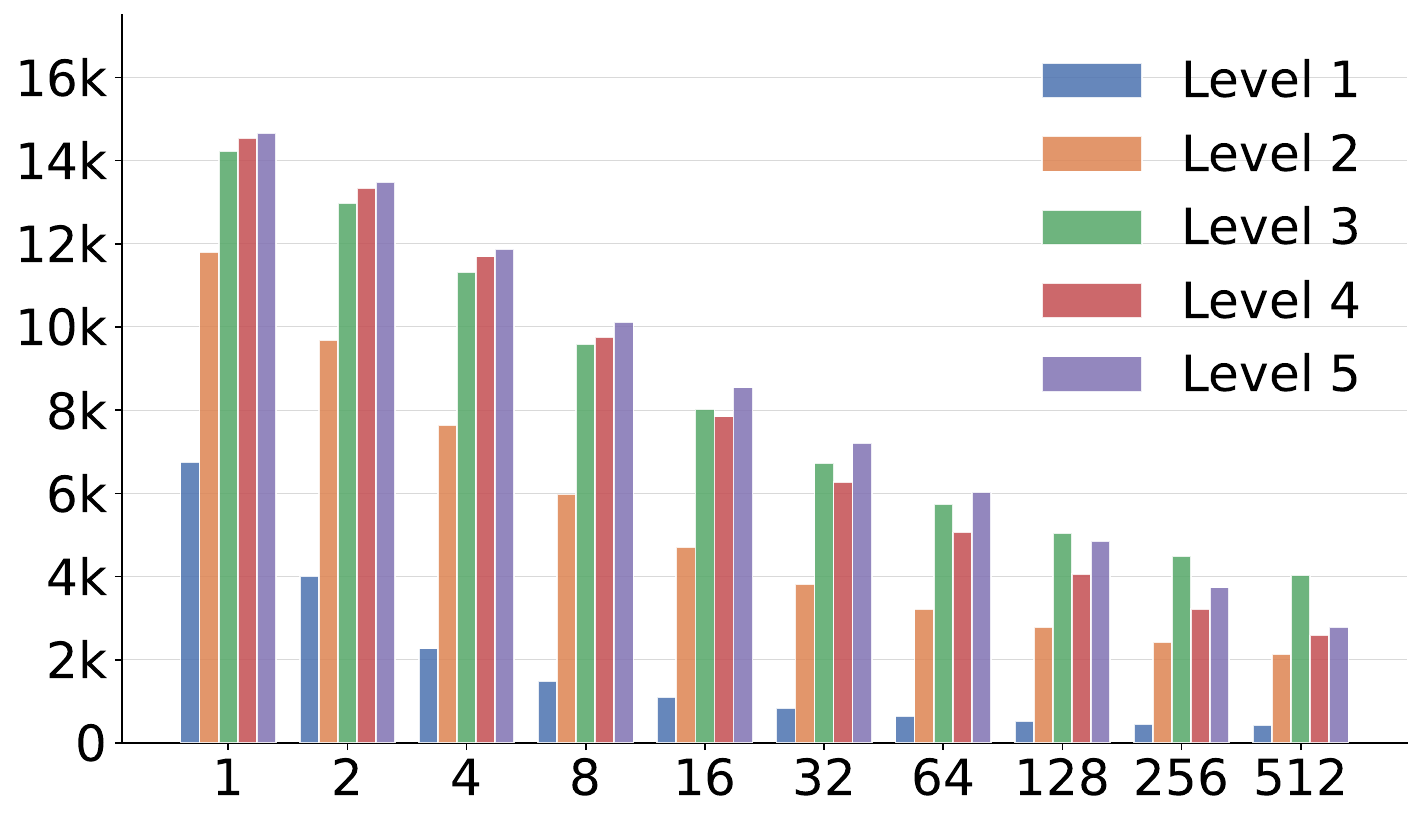}
            \caption{High-temperature}
        \end{subfigure}
        \hfill
        \begin{subfigure}[t]{0.48\linewidth}
            \centering
            \includegraphics[width=\linewidth]{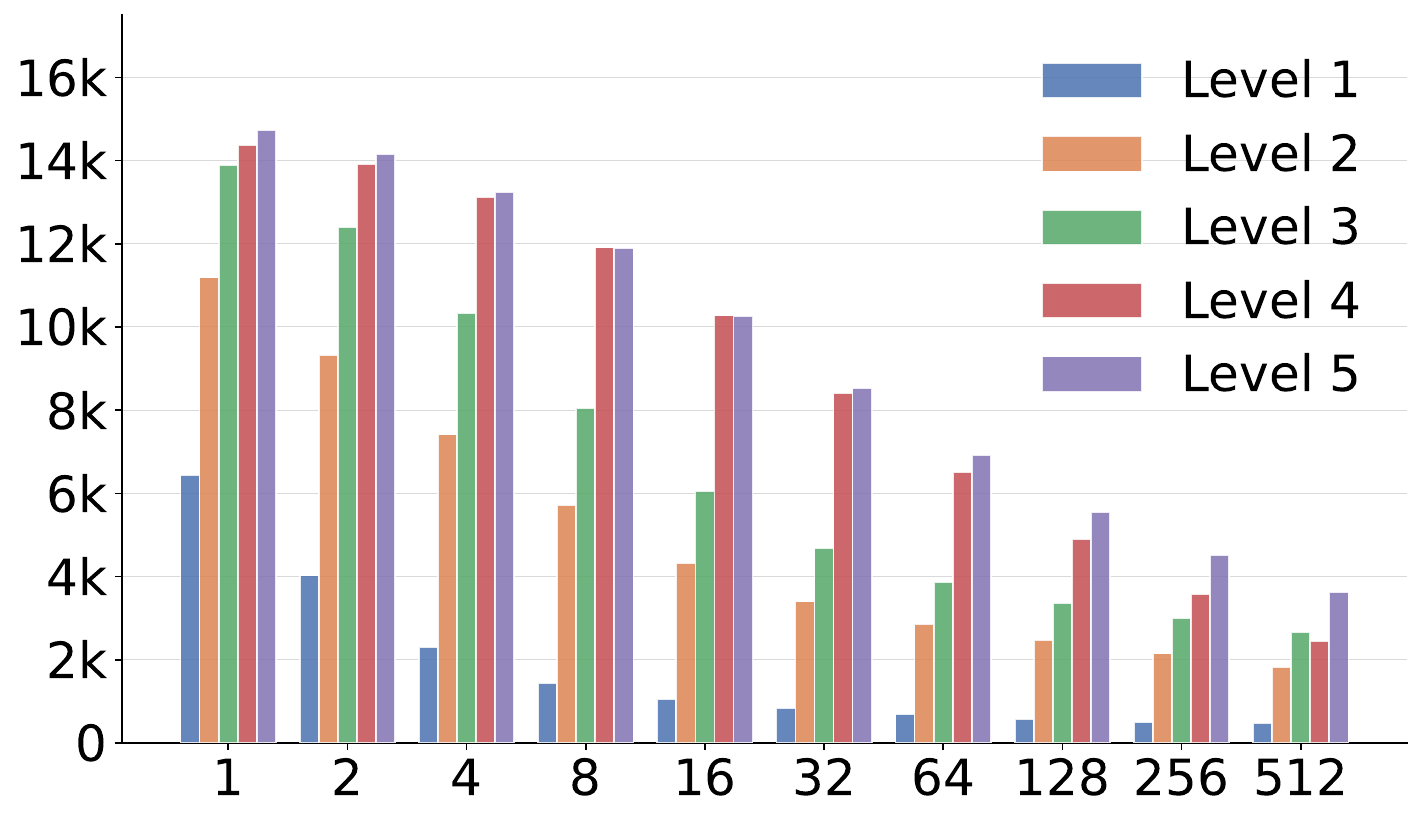}
            \caption{Constrained stochastic}
        \end{subfigure}
    \end{minipage}
    \hfill
    \begin{minipage}[c]{0.48\textwidth}
        \centering
        \begin{subfigure}[c]{\linewidth}
            \centering
            \includegraphics[width=\linewidth]{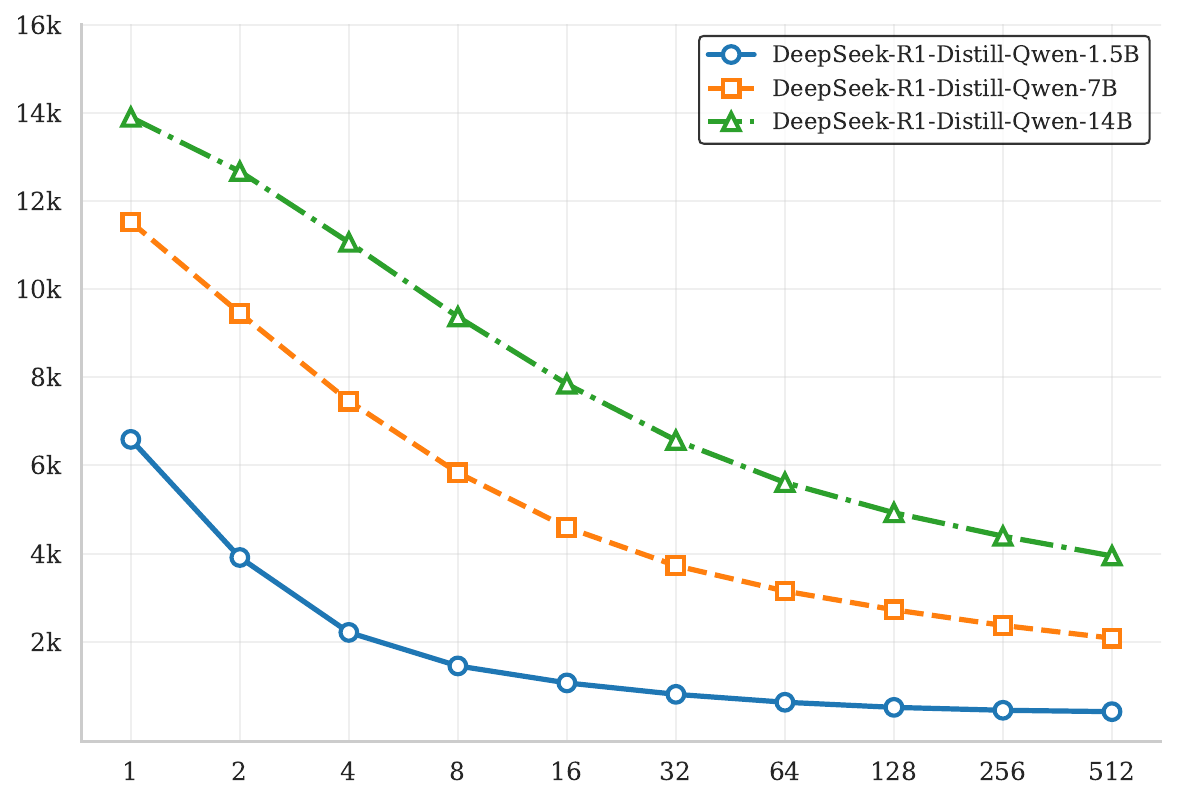}
            \caption{Scaling effect under varying difficulty}
        \end{subfigure}
    \end{minipage}

    \caption{
    Expected token length of the shortest correct reasoning path versus sample number $k$.
    (a–d) Different sampling strategies.
    (e) Model scaling under varying difficulty levels.
    }
    \label{fig:MSL-Combined}
\end{figure*}

\subsection{Learning-based Compression} Unlike inference-time interventions, training-based approaches reconstruct the reasoning process to eliminate redundancy. These approaches can be broadly categorized into symbolic and latent compression. One branch utilizes Supervised Fine-Tuning and distillation for symbolic compression. Methods like DRP and TokenSkip~\cite{jiang2025drp, xia2502tokenskip} prune low-value tokens, while Munkhbat et al.~\cite{munkhbat2025self} employ self-training to bias models toward concise trajectories. Conversely, latent approaches shift from discrete text to continuous representations. Works such as Token Assorted and LightThinker~\cite{hao2024training, su2025token, zhang2025lightthinker, tack2025llm} map CoT into latent spaces or high-level concepts, preserving reasoning capabilities while bypassing verbose generation.

RL provides another mechanism to align reasoning length with task complexity. A primary stream incorporates length penalties or token pruning into the reward function, as seen in L1, ConciseRL, and ThinkPrune~\cite{aggarwal2025l1, dumitru2025conciserl, liu2025learn, cheng2025optimizing, luo2025o1, hou2025thinkprune}. Beyond static penalties, \textit{adaptive computation} frameworks like DAST and AutoThink~\cite{shen2025dast, tu2025learning, zhang2025adaptthink, han2025your} dynamically allocate inference budgets based on problem difficulty. Furthermore, methods like ARM~\cite{wu2025arm, yi2025shorterbetter} optimize the reasoning format itself to ensure minimal sufficiency.

Despite these advancements, existing works fail to explicitly characterize the minimal reasoning length—or compression limit—for specific tasks. Addressing this, we profile the CoT compression limits of base models and propose \textbf{TRiMS}, a novel RL-based method designed to achieve optimal compression without compromising reasoning integrity.


\section{Minimal Sufficient Length}
\label{sec:msl}

\subsection{Definition of Minimal Sufficient Length}

Let $\mathcal{O}_k=\{\pi_i\}_{i=1}^k$ denote the set of $k$ trajectories generated by the model. We define a constant penalty length $L_{\max} = 16384$.

For the $i$-th trajectory $\pi_i$, we define its \textit{effective length} $l_i$ based on its correctness:
\[
l_i = 
\begin{cases} 
\operatorname{len}(\pi_i), & \text{if } \pi_i \text{ is correct}, \\
L_{\max}, & \text{otherwise}.
\end{cases}
\]

Let $\mathcal{L}_k = \{l_1, l_2, \dots, l_k\}$ be the set of effective lengths corresponding to the first $k$ samples. The shortest correct path length at step $k$, denoted as $L_k$, is defined as the minimum value in this set:
\[
L_k = \min \{ l_j \mid 1 \le j \le k \}.
\]

Correspondingly, the recursive form simplifies to:
\[
L_k = \min(L_{k-1}, l_k).
\]

As $k \to \infty$, the sequence $\{L_k\}$ is monotonically non-increasing and bounded below by $0$. Thus, the limit exists:
\[
\mathcal{A} = \lim_{k\to \infty} L_k.
\]
We refer to $\mathcal{A}$ as the \textbf{Minimal Sufficient Length}.

\subsection{Existence of Minimal Sufficient Length Under Different Sampling Strategies}
\label{sec:MSL-Stategy}

\begin{figure*}[!t]
    \centering
    \includegraphics[width=\linewidth]{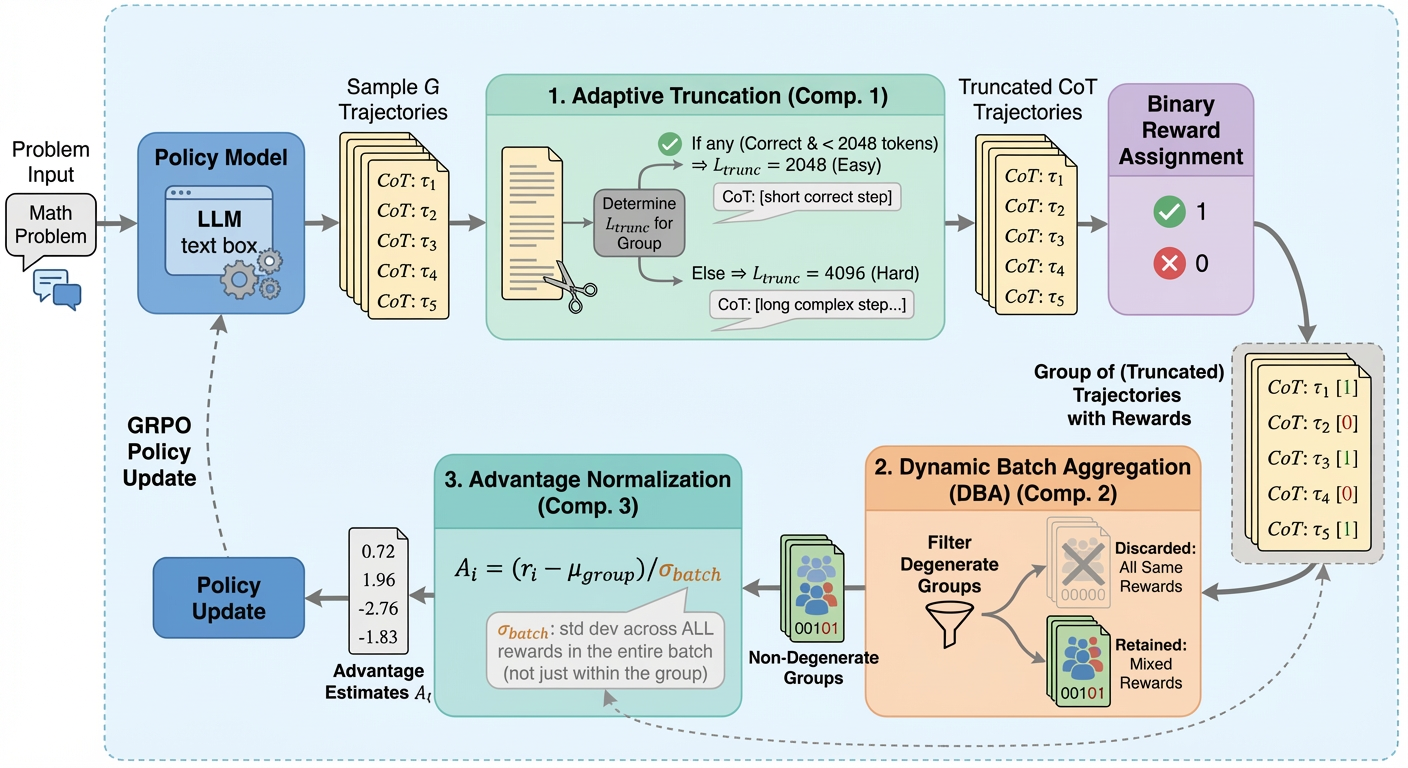}
    \caption{Overview of the TRiMS framework}
    \label{fig:Pipeline}
\end{figure*}

As illustrated in Fig.~\ref{fig:MSL-Combined}(a-d), to evaluate the robustness of MSL under different sampling configurations, we consider four sampling strategies: low-temperature, moderate-temperature, high-temperature, and constrained stochastic sampling. Detailed experimental settings are provided in the Appendix~\ref{Appendix:A.1}. The results demonstrate that, across all sampling regimes and difficulty levels, the observed trends remain consistent with those reported in the introduction: as the number of samples increases, the shortest correct reasoning path produced by the model decreases substantially and gradually converges to a stable value.

\subsection{Existence of Minimal Sufficient Length Across Model Scales}
\label{sec:MSL-Scale}

As shown in Fig.~\ref{fig:MSL-Combined}(e), to verify whether MSL is robust across different model scales, we evaluate three model sizes—1.5B, 7B, and 14B—under Difficulty Level 3. Setup is consistent with Sec.~\ref{sec:MSL-Stategy}. The results demonstrate a consistent trend across all scales: as the number of samples increases, the length of the shortest correct trajectory decreases monotonically and converges to a stable value. Moreover, smaller models yield shorter initial paths and lower convergence.


\section{TRiMS}
\label{sec:trims}
\subsection{Rationale}

\paragraph{Why Use RL}

While MSL provides a theoretical basis for CoT compression, its accurate estimation is affected by the model's generation distribution. Existing approaches mainly fall into supervised fine-tuning and reinforcement learning. Supervised fine-tuning may introduce distribution shifts, altering generation probabilities and leading to inaccurate MSL estimation. In contrast, reinforcement learning adjusts the model's generation probabilities within its original capability, making it more likely to produce correct trajectories with lengths close to the MSL \cite{yue2025does}. Specifically, we adopt the GRPO method to efficiently explore multiple sampled trajectories while satisfying the requirement of extensive sampling for accurate MSL estimation.

\begin{figure}[!t]
    \centering
    \includegraphics[width=\linewidth]{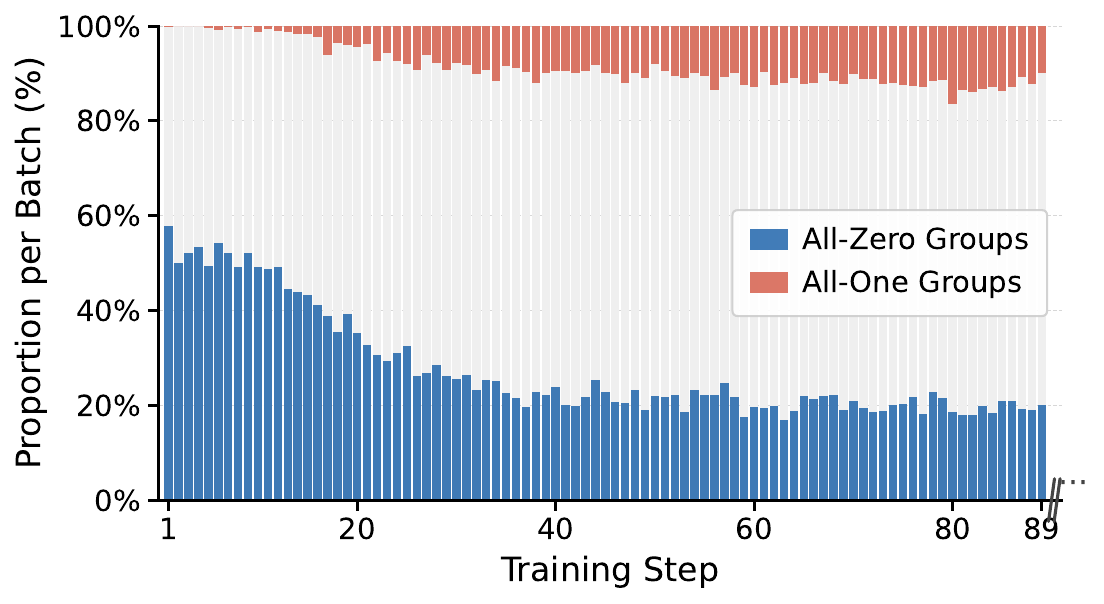}
    \caption{Percentage of degenerate groups across training steps}
    \label{fig:Method-DBA}
\end{figure}

\paragraph{Why Use Truncation}

Accurate MSL estimation requires sampling a large number of trajectories per problem (e.g., 256–512), which is computationally expensive. Experiments (Figs.~\ref{fig:introduction} and~\ref{fig:MSL-Combined}) show that, across various reasoning configurations, the shortest correct trajectory typically does not exceed 4096 tokens. Therefore, we adopt a truncation strategy at 4096 tokens, reducing computational cost while serving as a preliminary estimate of the MSL.



\subsection{Dynamic Batch Aggregation}

As shown in Fig.~\ref{fig:Method-DBA}, we track the proportion of \textit{all-zero groups} (bottom bars) and \textit{all-one groups} (top bars) across training steps. All-zero groups dominate early training ($>$50\%), gradually declining as optimization proceeds, while all-one groups persist throughout. Both types collapse under group-relative normalization, yielding zero advantages and destabilizing early-stage training. We therefore propose Dynamic Batch Aggregation, which filters out such degenerate groups before advantage estimation to ensure informative gradient signals.

\begin{figure}[!t]
    \centering
    \includegraphics[width=\linewidth]{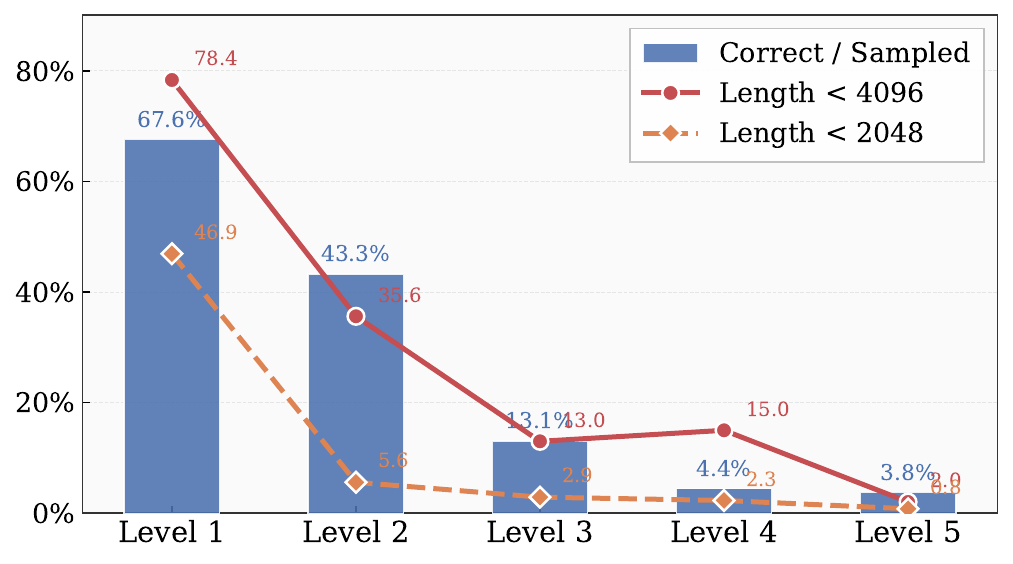}
    \caption{Proportion of correct answers distribution across different length thresholds and difficulty levels.}
    \label{fig:Method-truncation}
\end{figure}

\subsection{Adaptive Truncation}

While offline experiments show sufficient sampling captures correct trajectories under 4096 tokens, fixed truncation remains suboptimal across difficulty levels. Fig.~\ref{fig:Method-truncation} illustrates that easy problems typically terminate within 2048 tokens, unlike harder tasks. To address this, we propose a two-stage adaptive truncation strategy.

Let a group of sampled trajectories be \(Y = \{y_1, \dots, y_G\}\), where \(|y_i|\) denotes the length of trajectory \(y_i\) and \(\mathbf{1}_{\text{correct}}(y_i)\) indicates its correctness. The truncation length \(L_{\text{trunc}}\) is defined as:

\[
L_{\text{trunc}}(Y) =
\begin{cases}
2048, & \text{if } \exists\, y_i \in Y,\quad |y_i| < 2048, \\[-1mm]
           & \mathbf{1}_{\text{correct}}(y_i) = 1, \\[1mm]
4096, & \text{otherwise.}
\end{cases}
\]

This strategy balances computational efficiency with reliable coverage of minimal-length correct trajectories.

\begin{table*}[!t]
\centering
\caption{Comparison of PASS@1 accuracy, average output length (Token), and Intelligence Per Token (IPT) between the TRiMS model and baseline models across all benchmark datasets.}
\label{tab:Main-Results}
\resizebox{\textwidth}{!}{%
\begin{tabular}{lccccccccccccc}
\toprule
\multirow{2}{*}{\textbf{Method}} 
& \multicolumn{2}{c}{\textbf{AMC23}} 
& \multicolumn{2}{c}{\textbf{AIME24}} 
& \multicolumn{2}{c}{\textbf{MATH-500}} 
& \multicolumn{2}{c}{\textbf{Minerva}} 
& \multicolumn{2}{c}{\textbf{Olympiad}} 
& \multicolumn{3}{c}{\textbf{Avg}} \\
\cmidrule(lr){2-3} 
\cmidrule(lr){4-5} 
\cmidrule(lr){6-7} 
\cmidrule(lr){8-9} 
\cmidrule(lr){10-11} 
\cmidrule(lr){12-14}
& \textbf{Acc} 
& \textbf{Token} 
& \textbf{Acc} 
& \textbf{Token} 
& \textbf{Acc} 
& \textbf{Token} 
& \textbf{Acc} 
& \textbf{Token} 
& \textbf{Acc} 
& \textbf{Token} 
& $\Delta$\textbf{Acc (\%)} 
& $\Delta$\textbf{Token (\%)} 
& \textbf{IPT} \\
\midrule

\multicolumn{14}{l}{\textbf{DeepSeek-R1-Distill-Qwen-1.5B}} \\
Vanilla    & 61.1          & 8458.9         & 29.2          & 12303.4        & 82.3          & 4619.2        & 25.7          & 6284.1        & 41.0          & 8994.3        & --            & --             & 7.2 \\
LC-R1      & 56.5          & 3862.9         & 19.3          & 6737.5         & 71.2          & 2080.4        & 26.5          & 3827.3        & 37.9          & 3950.1        & -11.9         & -49.9          & 13.6 \\
Laser      & 60.2          & 4714.9         & \textbf{30.4} & \underline{5713.1} & \underline{84.1} & 1974.1   & \textbf{33.1} & 3218.3        & \underline{44.3} & 5645.9     & \textbf{+8.4} & -48.2          & 15.8 \\
AutoThink  & 57.8          & 4244.9         & \underline{30.0} & 8984.4      & \textbf{85.6} & 2111.5        & \underline{29.0} & 3031.5     & \textbf{44.7} & 5072.6        & +1.9          & -45.3          & 14.9 \\
AdaptThink & \underline{63.3} & \underline{3415.2} & 29.6    & 7268.0         & 82.7          & \underline{1821.9} & 25.5     & 1876.8        & 43.6          & 3970.8        & +2.2          & -57.4          & \underline{18.5} \\
JET        & 57.8          & 3563.4         & 29.4          & 6121.9         & 72.0          & 1909.6        & 23.9          & \underline{1839.1} & 43.7      & \underline{3665.8} & -3.5     & \underline{-60.1}  & 16.7 \\
TRiMS      & \textbf{71.1} & \textbf{1382.9} & 29.8         & \textbf{2196.5} & 83.4         & \textbf{836.3} & 28.3         & \textbf{958.3} & 42.7         & \textbf{1307.0} & \underline{+6.8} & \textbf{-83.5} & \textbf{45.1} \\

\midrule
\multicolumn{14}{l}{\textbf{DeepSeek-R1-Distill-Qwen-7B}} \\
Vanilla    & 79.5          & 6668.7         & 51.7          & 10688.6        & 91.0          & 3682.5        & 34.2          & 4907.5        & 54.8          & 7393.4        & --            & --             & 11.2 \\
LC-R1      & 74.2          & 3127.9         & 46.0          & 5859.9         & 84.4          & \underline{1381.7} & \underline{42.8} & \underline{1298.8} & 53.9    & 3239.8        & -0.3          & \underline{-58.1}  & \underline{28.4} \\
Laser      & \textbf{82.8} & \underline{2882.9} & 51.7       & \underline{4860.9} & \underline{91.8} & 1648.3    & \textbf{43.9} & 1826.0        & \textbf{60.5} & \underline{3061.9} & \textbf{+8.8} & -57.6       & 27.8 \\
AutoThink  & \underline{81.9} & 4616.0      & \underline{53.3} & 6909.5      & \textbf{92.0} & 2378.2        & 35.3          & 2512.2        & \underline{56.1} & 5052.4     & +2.6          & -36.4          & 17.9 \\
AdaptThink & 79.5          & 4444.9         & \textbf{53.8} & 8738.8         & 90.4          & 1871.4        & 34.9          & 2362.0        & 53.6          & 5658.7        & \underline{+4.7} & -35.2       & 19.3 \\
JET        & 79.5          & 3979.2         & 52.9          & 7863.6         & 89.2          & 1935.7        & 37.5          & 1786.1        & 55.1          & 4653.3        & +2.1          & -43.0          & 21.1 \\
TRiMS      & \underline{81.9} & \textbf{1088.3} & 49.6       & \textbf{1531.2} & 91.4         & \textbf{712.4} & 39.0         & \textbf{696.8} & 54.4         & \textbf{1104.1} & +2.5        & \textbf{-84.2} & \textbf{68.2} \\

\bottomrule
\end{tabular}%
}
\end{table*}

\begin{figure}[!ht]
    \centering
    \includegraphics[width=\linewidth]{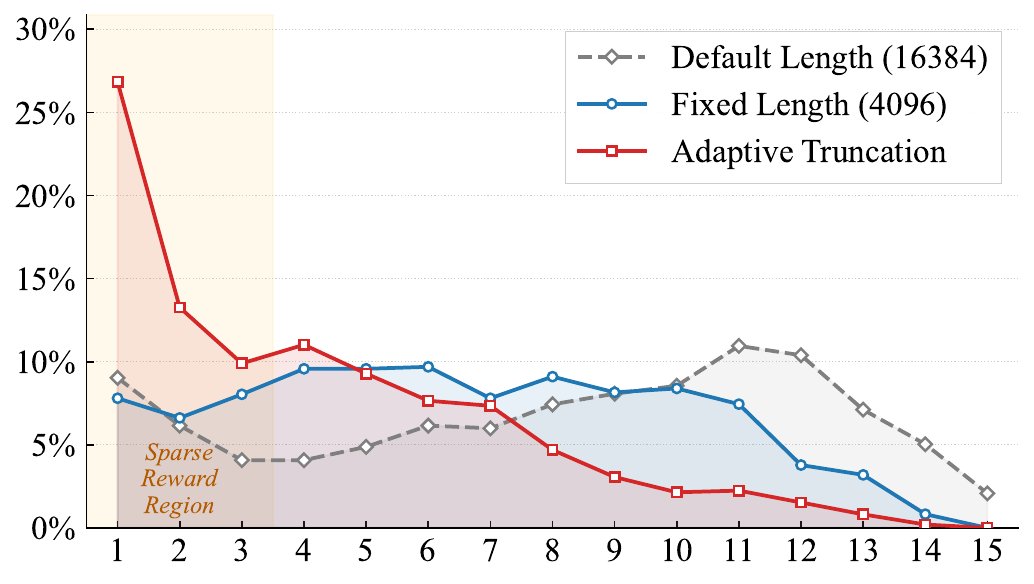}
    \caption{Distribution of correct responses per group ($k$) under different truncation strategies ($G=16$).}
    \label{fig:Method-std}
\end{figure}

\subsection{Advantage Normalization under Reward Sparsity}
\label{method:Adv}

GRPO typically normalizes rewards within each response group: 
$\hat{A}_{i}^{\text{group}} = (R_{i} - \mu_{R}^{\text{group}}) / \sigma_{R}^{\text{group}}$. 
For binary rewards $r_i \in \{0,1\}$ with $k$ correct responses out of $G$, the group standard deviation $\sigma_{R}^{\text{group}} = \sqrt{k(G-k)}/G$ ensures a constant gradient energy $\sum_i (\hat{A}_i)^2 = G$.

However, as shown in Fig.~\ref{fig:Method-std}, aggressive length truncation causes severe reward sparsity ($k \ll G$), where a small $\sigma_{R}^{\text{group}}$ (e.g., $k=1$) excessively amplifies advantages. This leads to high-variance gradients and training instability, a phenomenon also observed in REINFORCE++~\cite{hu2025reinforce++}. To stabilize training, we replace the group-level denominator with a batch-level one:
\begin{equation}
\hat{A}_{i}^{\text{batch}} = \frac{R_{i} - \mu_{R}^{\text{group}}}{\sigma_{R}^{\text{batch}}},
\end{equation}
where $\sigma_{R}^{\text{batch}}$ is computed across all rewards in the batch. This modification scales each group's gradient energy by $(\sigma_{R}^{\text{group}})^2 / (\sigma_{R}^{\text{batch}})^2$, preventing rare, sparse-reward groups from producing disproportionately large gradients and ensuring cross-group comparability.

\section{Experiments}

\subsection{Experiments Setup}
\label{sec:setup}

\paragraph{Model Selection}
We employ \textbf{DeepSeek-R1-Distill-Qwen-1.5B} and \textbf{DeepSeek-R1-Distill-Qwen-7B} as the foundation models for our experiments~\cite{deepseekai2025deepseekr1incentivizingreasoningcapability}. 

\paragraph{Training Implementation}
The training process is conducted on the DeepScaleR dataset, as detailed in Appendix~\ref{Appendix:B.2}, using the {verl}~\cite{sheng2024hybridflow} framework. The training hyperparameters are provided in Appendix~\ref{Appendix:B.1}.

\paragraph{Evaluation Protocol}
We evaluate models on five benchmark datasets: AIME24, AMC23, MATH-500, Minerva, and OlympiadBench. Detailed descriptions of these datasets are provided in Appendix~\ref{Appendix:B.2}. We report \textbf{Acc} and \textbf{Token} as primary metrics, along with the dataset-level \textbf{$\Delta$}. To jointly evaluate accuracy and cost, we define \textbf{IPT} as {Intelligence Per Token}. See Appendix~\ref{Appendix:B.3} for details.

\paragraph{Baseline}
We compare \ourMethod with the following representative methods for efficient reasoning:
LC-R1, Laser, AutoThink, AdaptThink, JET.
Detailed descriptions are provided in the Appendix~\ref{Appendix:B.4}.

\begin{figure*}[!t]
    \centering
    \begin{subfigure}[b]{0.19\textwidth}
        \includegraphics[width=\textwidth]{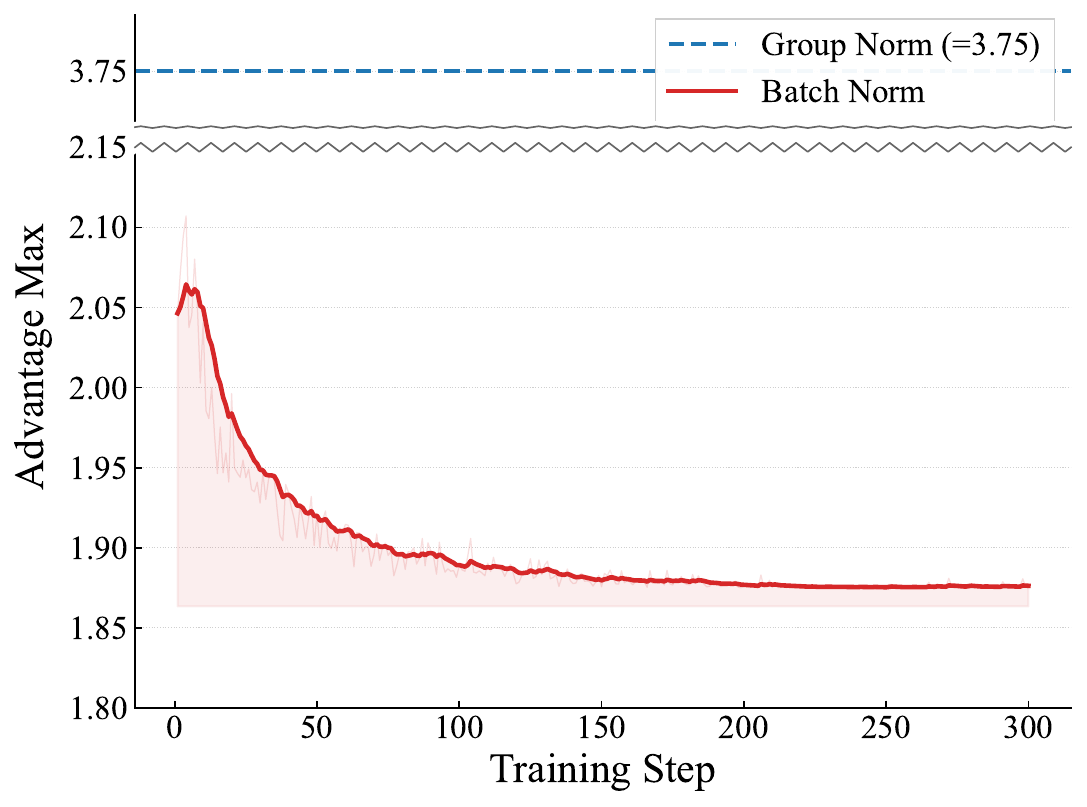}
        \caption{Adv.\ Max}
        \label{fig:adv-norm-max}
    \end{subfigure}
    \hfill
    \begin{subfigure}[b]{0.19\textwidth}
        \includegraphics[width=\textwidth]{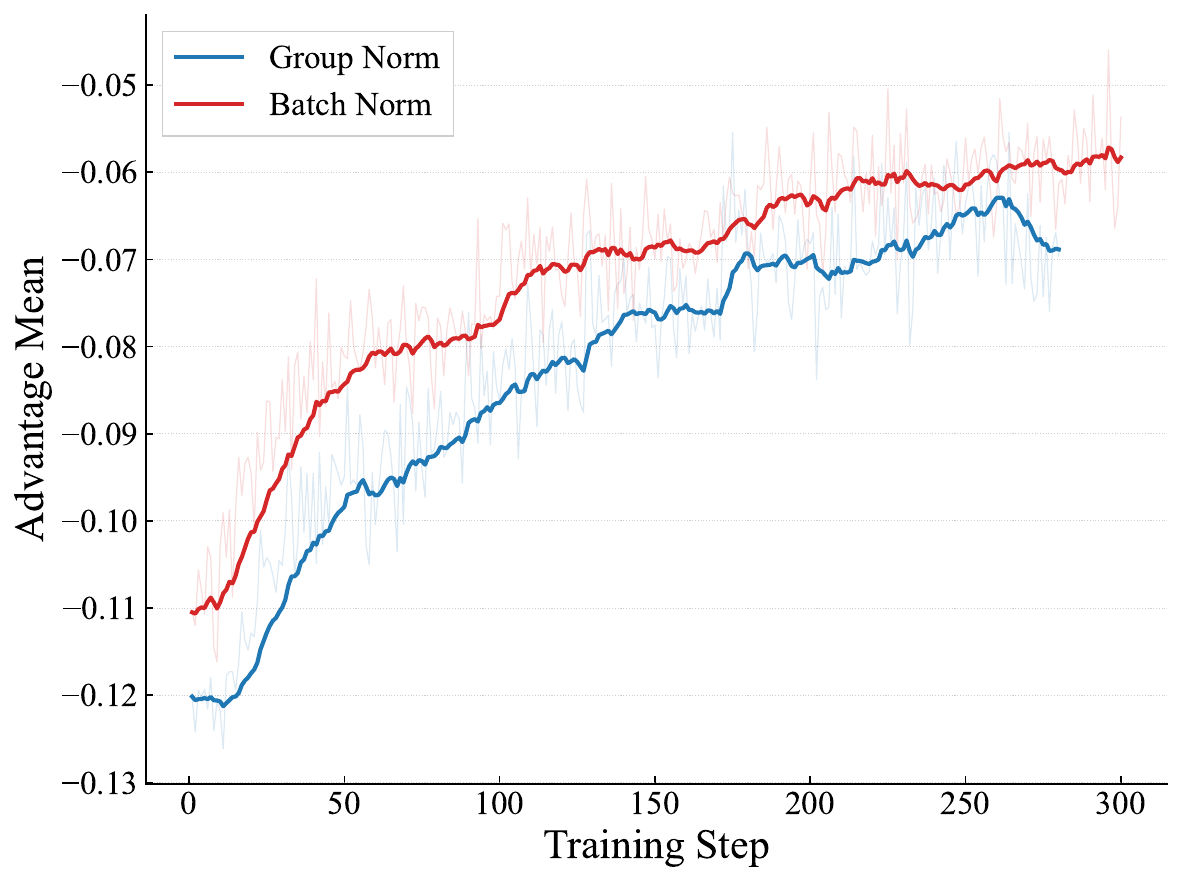}
        \caption{Adv.\ Mean}
        \label{fig:adv-norm-mean}
    \end{subfigure}
    \hfill
    \begin{subfigure}[b]{0.19\textwidth}
        \includegraphics[width=\textwidth]{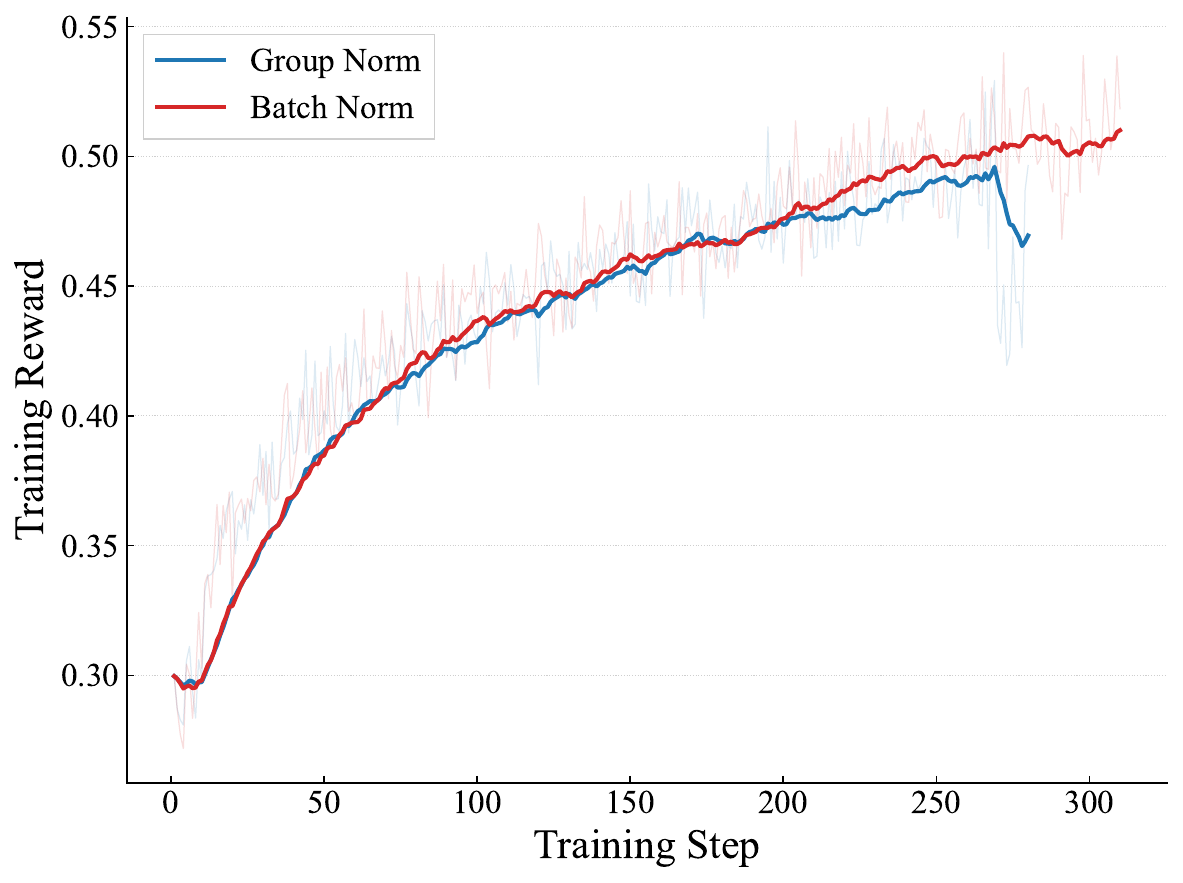}
        \caption{Training Reward}
        \label{fig:adv-norm-reward}
    \end{subfigure}
    \hfill
    \begin{subfigure}[b]{0.19\textwidth}
        \includegraphics[width=\textwidth]{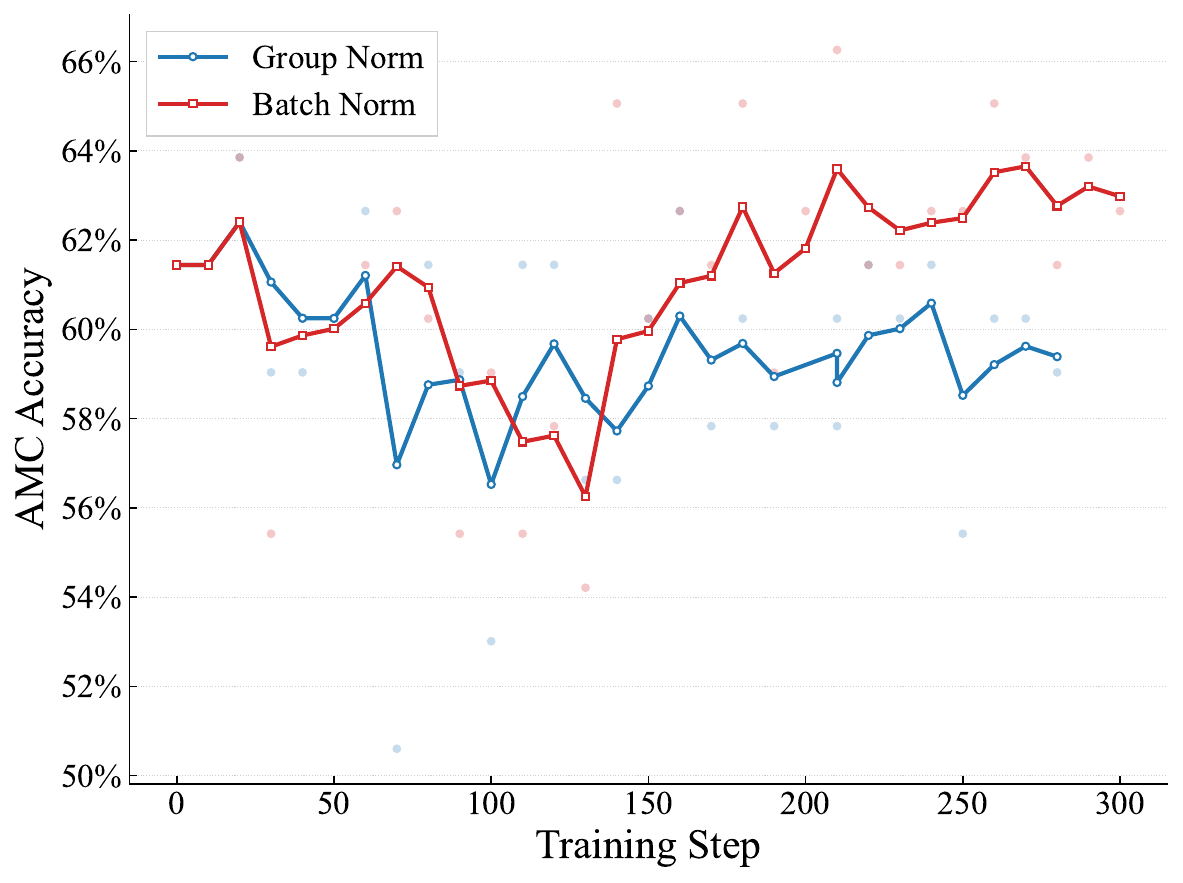}
        \caption{AMC Accuracy}
        \label{fig:adv-norm-acc}
    \end{subfigure}
    \hfill
    \begin{subfigure}[b]{0.19\textwidth}
        \includegraphics[width=\textwidth]{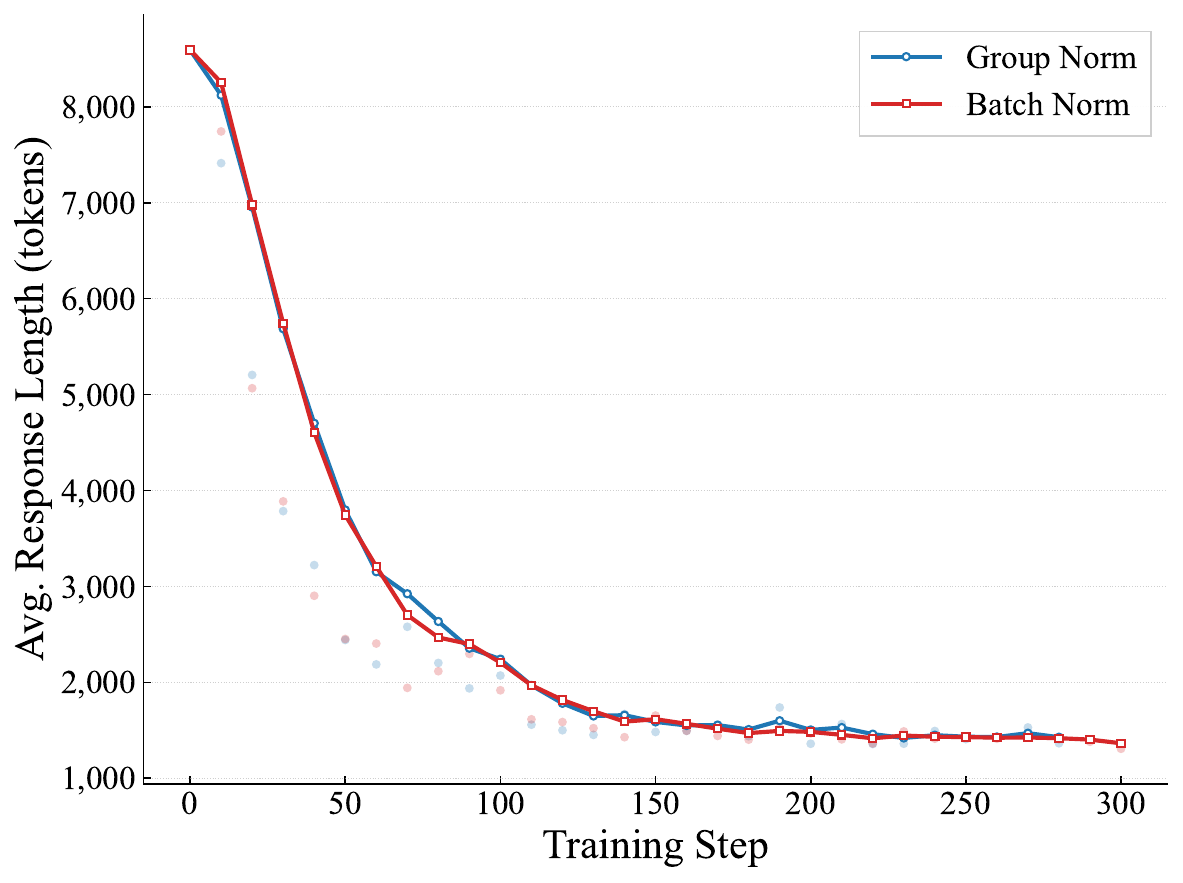}
        \caption{Response Length}
        \label{fig:adv-norm-length}
    \end{subfigure}
    \caption{Group-level vs.\ batch-level advantage normalization throughout training.}
    \label{fig:adv-norm}
\end{figure*}

\subsection{Main Results}

Table~\ref{tab:Main-Results} presents the full comparison across five benchmarks on both the 1.5B and 7B scales. We highlight the following findings:

\paragraph{TRiMS achieves the highest compression with accuracy gains.}
On the 1.5B model, TRiMS reduces token consumption by \textbf{83.5\%} while improving accuracy by \textbf{+6.8\%}—the only method achieving both substantial compression and accuracy improvement. The best competing method in compression, JET, attains $-60.1\%$ tokens but suffers $-3.5\%$ accuracy loss. Results on the 7B model ($-84.2\%$ tokens, $+2.5\%$ accuracy) confirm robustness across scales.

\paragraph{TRiMS dominates in Intelligence Per Token.}
TRiMS achieves an IPT of \textbf{45.1} on 1.5B ($6.3\times$ the Vanilla baseline, $2.4\times$ the best competitor AdaptThink) and \textbf{68.2} on 7B (vs. LC-R1 28.4), confirming that the compression reflects genuinely more efficient reasoning rather than quality sacrifice.

\paragraph{Comparison with baselines.}
LC-R1's aggressive length penalty causes severe accuracy degradation ($-11.9\%$ on 1.5B), illustrating risks of static compression objectives. Laser and AdaptThink strike a better balance but their compression ($-35.2\%$ to $-57.6\%$) remains below TRiMS. JET provides $-60.1\%$ on 1.5B, yet TRiMS doubles this without accuracy loss. These results validate our thesis: by optimizing toward the MSL lower bound, TRiMS unlocks a compression regime inaccessible to methods lacking a principled characterization of minimal necessary reasoning length.

\begin{figure}[!t]
    \centering
    \includegraphics[width=\linewidth]{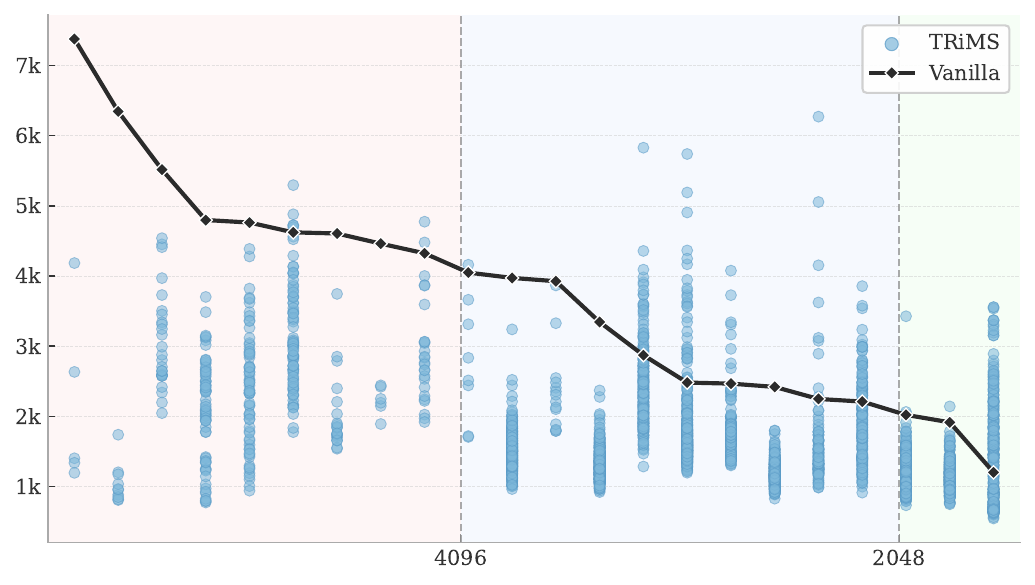}
    \caption{Per-problem generation lengths on AIME24. TRiMS (blue dots) is compared against the empirically estimated MSL (black curve, SCPT@k with k=512), sorted by descending MSL estimate.}
    \label{fig:distance_msl}
\end{figure}

\subsection{Distance to MSL}
\label{sec:distance_msl}


To validate that TRiMS steers generation toward the MSL lower bound, we compare per-problem generation lengths of TRiMS against the empirically estimated MSL (approximated via SCPT@k with k=512, see Appendix~\ref{Appendix:A.2}) on AIME24. Problems are sorted by descending MSL estimate, with dashed lines marking the 4096- and 2048-token thresholds. As shown in Fig.~\ref{fig:distance_msl}, TRiMS consistently approaches the estimated MSL when it exceeds 2048 tokens, with the largest reductions in the high-length regime. However, for problems whose estimated MSL already falls below 2048 tokens, TRiMS does not further compress, as the reward signal is not explicitly designed to incentivize compression in this regime.

\subsection{Impact of Advantage Normalization}

We compare \textbf{Group Norm} (standard GRPO) with \textbf{Batch Norm} (ours) on CoT compression, keeping all other hyperparameters identical.

As shown in Fig.~\ref{fig:adv-norm}, Group Norm yields an advantage maximum of 3.75, while Batch Norm halves it to~1.88. Sparse-reward groups have small $\sigma_{\text{group}}$, inflating per-sample advantages. During multi-epoch PPO updates, these inflated values push $\rho$ away from~1 rapidly, triggering excessive clipping---an {amplify-then-truncate} effect that wastes gradient signal. Batch Norm avoids this with a shared, more stable $\sigma^{\text{batch}}$. This stability gap translates into performance: both methods follow similar reward trajectories, but Group Norm suffers a sharp drop near step~280. On AMC, Group Norm accuracy falls below 5\% mid-training with response length rebounding to ${\sim}$900 tokens, whereas Batch Norm maintains accuracy above 10\%.

\subsection{Reasoning Trajectory Analysis}

If TRiMS successfully drives the model toward MSL, we expect two observable effects: (1)~systematic elimination of {redundant reasoning patterns}---such as repetitive backtracking and hesitation; and (2)~{difficulty-adaptive compression}, where compression scales naturally with problem complexity. We verify both hypotheses below.

\subsubsection{Lexical Repetition Analysis}
\label{sec:lexical}
\begin{figure}[!ht]
    \centering
    \includegraphics[width=\linewidth]{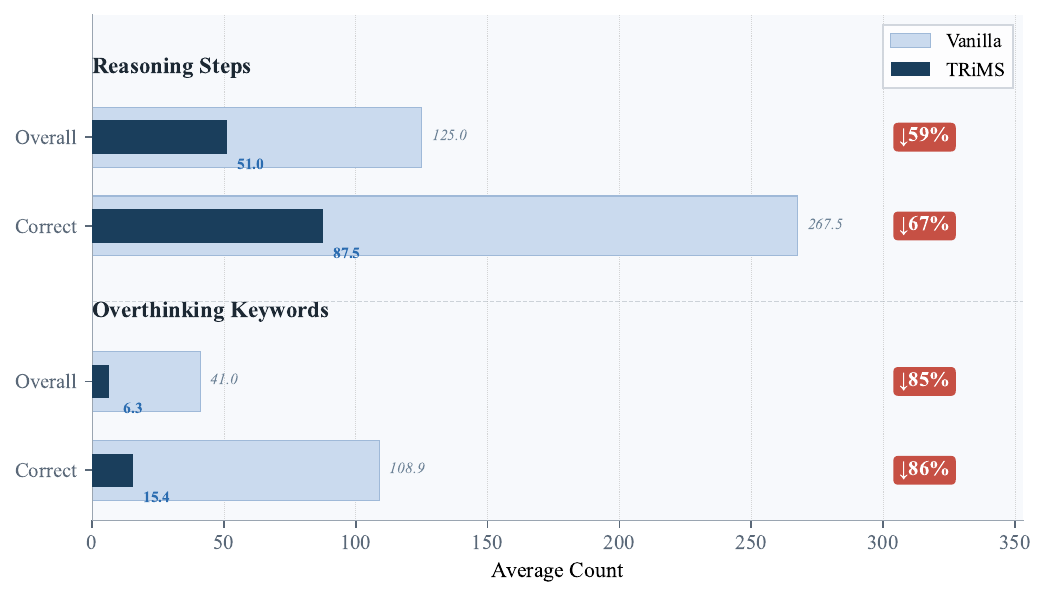}
    \caption{Reasoning trace statistics on AIME-24: Comparison of reasoning steps and overthinking keyword frequencies.}
    \label{fig:trajectory_lpa}
\end{figure}

We segment reasoning traces by double-newline delimiters and adopt the overthinking indicators from~\citet{lu2025retro}: \{\textit{But}, \textit{Wait}, \textit{Alternatively}, \textit{However}, \textit{Hmm}, \textit{Hmmm}, \textit{Not sure}, \textit{Going back}, \textit{Backtrack}, \textit{Trace back}, \textit{Another}\}---keywords signaling hesitation, self-correction, or others.

As shown in Fig.~\ref{fig:trajectory_lpa}, TRiMS substantially reduces both reasoning steps and overthinking keywords on AIME-24. Compared with Vanilla, TRiMS cuts reasoning steps by 59\% overall (125.0$\to$51.0) and 67\% on correct responses (267.5$\to$87.5). The reduction in overthinking keywords is even more pronounced: 85\% overall (41.0$\to$6.3) and 86\% on correct responses (108.9$\to$15.4). This disproportionate drop indicates that the eliminated tokens consist predominantly of repetitive backtracking and hesitation loops rather than productive reasoning. Combined with maintained or improved accuracy (Table~\ref{tab:Main-Results}), this confirms that TRiMS selectively suppresses redundant patterns while preserving the essential reasoning chain.

\subsubsection{Difficulty-Adaptive Token Analysis}
\label{sec:exp-difficulty}

\begin{figure}[!ht]
    \centering
    \includegraphics[width=\linewidth]{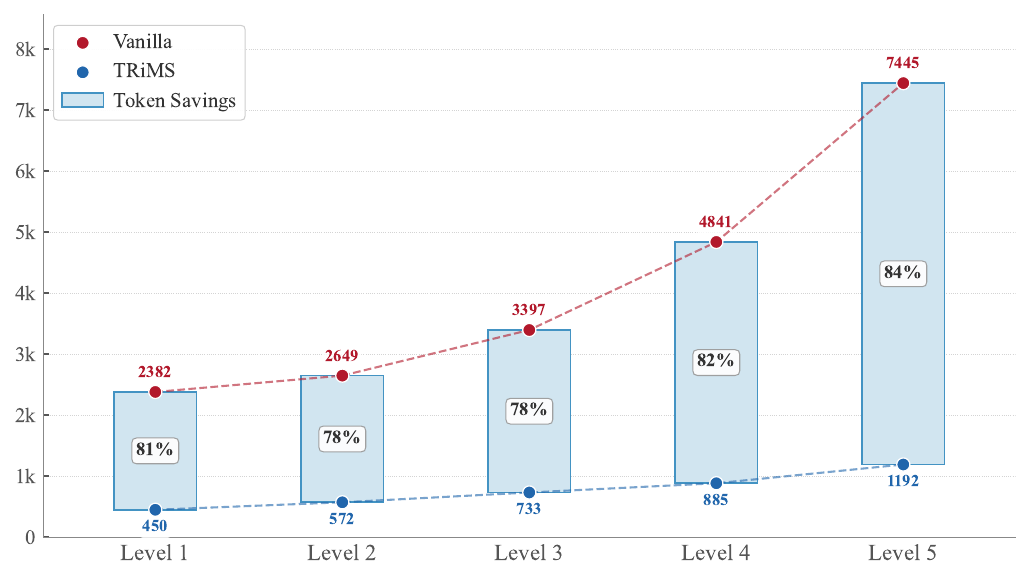}
    \caption{Average response length of TRiMS across five difficulty levels on MATH-500 (1.5B model).}
    \label{fig:trajectory_difficulty}
\end{figure}

We further examine whether TRiMS compression is difficulty-adaptive. Using the 1.5B model on MATH-500 with the same setup as Sec.~\ref{sec:setup}, we report average response length for both Vanilla and TRiMS across difficulty levels. Per-level accuracy is provided in Appendix~\ref{Appendix:D.1}.

As shown in Fig.~\ref{fig:trajectory_difficulty}, TRiMS achieves consistently high compression across all levels, with token savings ranging from 78\% (Level~2--3) to 84\% (Level~5). While the compression {ratio} remains stable, {absolute} savings scale with difficulty: TRiMS saves approximately 1{.}9k tokens on Level~1 but over 6{.}2k on Level~5. Notably, this proportional compression emerges from a uniform training objective without explicit difficulty conditioning, consistent with the MSL convergence property established in Sec.~\ref{sec:MSL-Stategy} and~\ref{sec:MSL-Scale}: harder problems demand longer reasoning chains and thus contain proportionally more compressible redundancy. Together with the lexical analysis above, these results confirm that TRiMS achieves difficulty-adaptive compression.

\subsection{Efficiency under Fixed Inference Budget}
\label{sec:efficiency}

\begin{figure}[!ht]
    \centering
    \includegraphics[width=\linewidth]{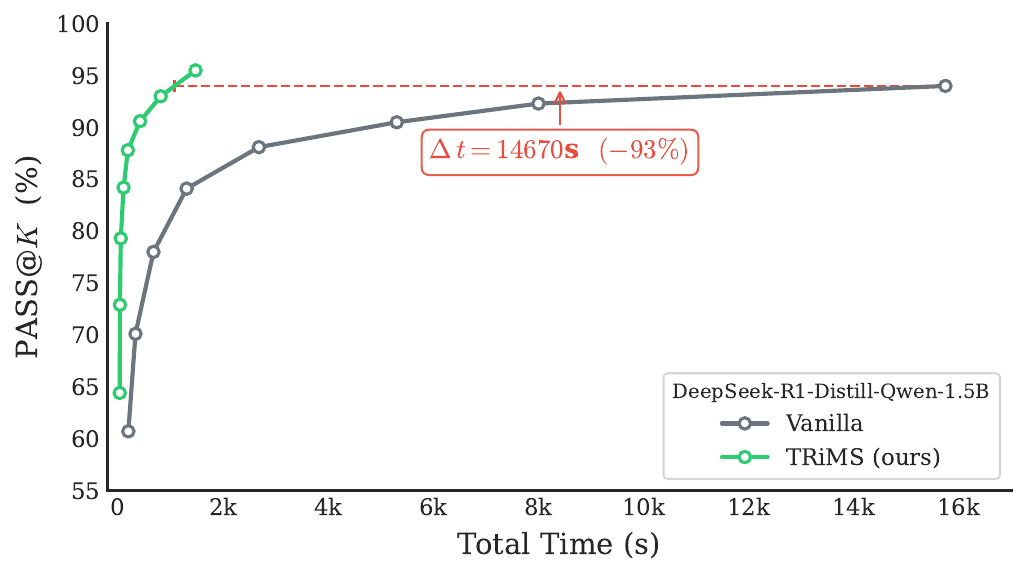}
    \caption{TRiMS fits more parallel trajectories within the same budget, substantially outperforming the Vanilla model on AMC23.}
    \label{fig:parallel_efficiency}
\end{figure}

Since TRiMS produces significantly shorter reasoning traces, it can generate far more parallel trajectories within the same inference time. We vary $n \in \{2, 4, 8, \ldots, 128\}$ and compare majority voting accuracy against total inference times. As shown in Fig.~\ref{fig:parallel_efficiency}, this advantage compounds with scale: TRiMS consistently surpasses the vanilla model at every comparable budget, achieving substantially higher accuracy by trading shorter chains for broader parallel exploration. Additional analysis are in Appendix~\ref{Appendix:D.2}.

\subsection{Generalization Analysis}

We evaluate TRiMS along two axes: (1) generalization to a stronger reasoning model, and (2) generalization to OOD benchmarks.

\subsubsection{Generalization to Stronger Models}
\begin{figure}[!ht]
    \centering
    \includegraphics[width=\linewidth]{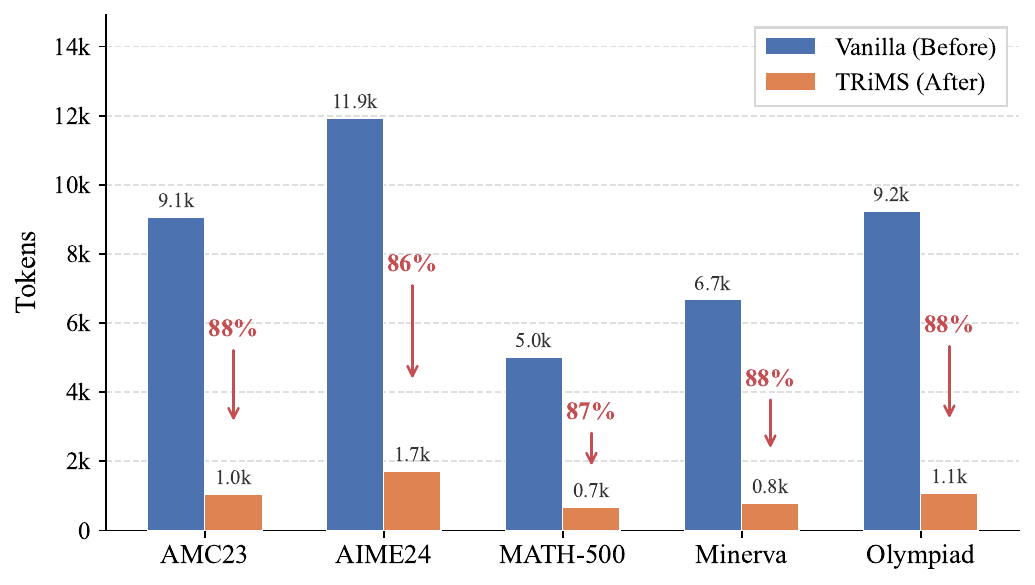}
    \caption{Qwen3-4B Tokens before and after training.}
    \label{fig:gen_sota}
\end{figure}

We apply TRiMS to Qwen3-4B to assess its transfer to a stronger, architecturally distinct backbone. Beyond the backbone change, we also upgrade the adaptive truncation strategy, with details provided in Appendix~\ref{Appendix:D.3}. The remaining setups follow Sec.~\ref{sec:setup}. As shown in Fig.~\ref{fig:gen_sota}, TRiMS achieves over 86\% token reduction across all benchmarks; additional results, including accuracy comparisons, are provided in Appendix~\ref{Appendix:D.4}. These observations suggest that the MSL-guided objective captures a structural property of reasoning trajectories that generalizes across model architecture and scales, rather than exploiting weaknesses specific to a particular model family.

\subsubsection{Generalization to Out-of-Distribution Scenarios}

\begin{table}[!ht]
\centering
\small
\setlength{\tabcolsep}{4pt}
\caption{OOD generalization results on MMLU.}
\label{tab:gen_ood}
\resizebox{\columnwidth}{!}{
\begin{tabular}{lcccccc}
\toprule
\multirow{2}{*}{Method} & \multicolumn{5}{c}{MMLU} \\
\cmidrule(lr){2-6}
 & Acc & Token & $\Delta$Acc & $\Delta$Token & IPT \\
\midrule
\multicolumn{6}{l}{\textit{DeepSeek-R1-Distill-Qwen-1.5B}} \\
Vanilla   & 38.1 & 1718.7  & -- & -- & 22.2 \\
TRiMS     & 48.2 & 388.1 & +26.5  & -77.4  & 124.2 \\
\midrule
\multicolumn{6}{l}{\textit{DeepSeek-R1-Distill-Qwen-7B}} \\
Vanilla   & 63.7 & 1158.5  & -- & -- & 55.0 \\
TRiMS     & 72.9 & 316.6  & +14.4  & -72.7  & 230.3 \\
\bottomrule
\end{tabular}
}
\end{table}

To test whether the efficiency learned by TRiMS transfers beyond its mathematical training distribution, we evaluate on MMLU. As shown in Table~\ref{tab:gen_ood}, TRiMS achieves a substantial accuracy improvement (+26.5\% on the 1.5B model) while reducing average response length by 77.4\%. Similar improvements are observed on the 7B model (+14.4\% accuracy and -72.7\% tokens). This indicates that TRiMS induces a generalizable tendency toward concise yet accurate reasoning, rather than merely shortening mathematical derivations.

\section{Conclusions}



We introduce Minimal Sufficient Length, formalizing the shortest correct reasoning path under a model's sampling distribution, and show through experiments (1.5B–14B) that it consistently exists and converges as a measurable lower bound for CoT compression. Building on this, we propose \ourMethod, an online RL framework steering models toward their MSL via adaptive truncation, dynamic batch aggregation, and batch-level advantage normalization—achieving over 80\% token reduction, +6.8\% accuracy improvement, and a 6.3× Intelligence-per-Token gain across backbones and OOD benchmarks, suggesting MSL captures a model-agnostic structural property of reasoning.
\clearpage

\section*{Limitations}
This work still has several limitations. (1) Due to computational constraints, our experiments were conducted only on models with 1.5B, 4B and 7B parameters. Although the results indicate that \ourMethod consistently yields stable and significant performance improvements across different scales, validation on larger models remains an important direction for future work. (2) Similar to most prior open-source studies, we primarily trained on mathematical data in order to obtain automatically verifiable reward signals. While evaluations on benchmarks such as MMLU suggest that \ourMethod exhibits a certain degree of out-of-distribution generalization, its applicability and performance could be further improved by incorporating more diverse, general-domain training data equipped with reliable verifiable rewards. (3) The proposed Minimal Sufficient Length currently characterizes only the shortest sufficient reasoning length induced by the model’s intrinsic sampling distribution under a fixed input prompt, and has not yet been extended to variations across different prompt formulations. Although prior work has empirically investigated the effect of prompt design on reasoning efficiency, we have not unified these factors within a single theoretical framework; a principled treatment of prompt-robust or conditional MSL remains an open direction for future research.(4) The adaptive truncation strategy adopted in TRiMS employs a coarse two-stage discrete approximation (2048/4096) to the MSL boundary. While effective in practice, this binary scheme does not fully exploit the continuous nature of MSL. Designing more fine-grained truncation schedules—such as per-problem adaptive thresholds or smooth length-based reward shaping that continuously guides generation toward MSL—remains an important direction for future work. Such strategies could yield tighter approximations to the theoretical MSL lower bound and potentially unlock further compression gains.

\bibliography{custom}

\clearpage
\appendix

\section{Supplementary Analysis of Minimal Sufficient Length}

\subsection{Sampling Protocol and Evaluation of Shortest Correct Path Tokens}
\label{Appendix:A.1} 
{

\begin{table}[t]
\centering
\caption{Sampling configurations used throughout the paper. 
The Moderate-temperature setting is adopted as the default unless otherwise specified.}
\label{tab:sampling_configs}

\resizebox{\columnwidth}{!}{%
\begin{tabular}{lcccc}
\toprule
Configuration & Temperature & top-$p$ & top-$k$ & Default \\
\midrule
Low-temperature   & 0.3 & 0.95 & $-1$  & No \\
Moderate-temperature & 0.6 & 0.95 & $-1$  & Yes \\
High-temperature  & 1.0 & 0.95 & $-1$  & No \\
Constrained Stochastic   & 0.6 & 1.0  & 100  & No \\
\bottomrule
\end{tabular}
}
\end{table}

In this section, we describe the experimental setup illustrated in Fig.~\ref{fig:introduction} (bottom).
All sampling configurations considered in this work are summarized in Table~\ref{tab:sampling_configs}.
Unless otherwise specified, we adopt the \emph{Moderate-temperature} configuration as the default setting, and all results reported in the introduction are based on this configuration.

For each problem, we perform 512 independent sampling runs.
Since only 16 samples are used for supervision during training, we retain as valid instances only those problems for which at least one correct answer is produced within the first 16 samples.
To mitigate potential distributional bias, we further enforce that each difficulty level contains at least 50 valid problems.
The difficulty stratification strictly follows the metadata provided by the DeepScaleR dataset.

To study the effect of sampling hyperparameters, we additionally evaluate the alternative configurations listed in Table~\ref{tab:sampling_configs}, including \emph{Low-temperature}, \emph{High-temperature}, and \emph{Constrained Stochastic}.
Except for the explicitly stated differences in the table, all other experimental settings are kept identical across these variants.
}

\subsection{Unbiased Estimation of Shortest Correct Path Tokens}
\label{Appendix:A.2}
{
To rigorously evaluate the coverage of LLM reasoning capabilities, we extend the $PASS@k$ metric \citep{chen2021evaluating, yue2025does} to measure the \textbf{Shortest Correct Path Tokens}. For a dataset $D$, the unbiased estimator for $PASS@k$ is given by:
\begin{equation}
    \text{PASS@}k := \mathbb{E}_{x_i \sim D} \left[ 1 - \frac{\binom{n - c_i}{k}}{\binom{n}{k}} \right]
\end{equation}
where $n$ is the total number of sampled trajectories and $c_i$ is the number of correct solutions for problem $x_i$.

Building upon this, we define the \textit{Shortest Correct Path Tokens} (SCPT@$k$) to estimate the expected minimum token count required to reach a correct solution within $k$ attempts. Given $n$ independent samples, let $\mathcal{L}_i$ be the set of lengths of correct trajectories for problem $x_i$. The algorithm for calculating the unbiased SCPT@$k$ is formalized in Algorithm~\ref{alg:msl}.

\begin{algorithm}[t]
\caption{Unbiased Estimation of Shortest Correct Path Tokens (SCPT@$k$)}
\label{alg:msl}
\begin{algorithmic}[1]
\Require For each problem $i$, sampled trajectories with correctness flags and lengths; 
        target set $\mathcal{K}=\{1,2,4,\dots,512\}$
\Ensure $\text{SCPT}[k]$ for each $k \in \mathcal{K}$

\ForAll{problems $i \in D$}
    \State Extract lengths of correct samples:
    \[
        \mathcal{C}_i = \{L_j \mid \text{sample } j \text{ is correct}\}, \quad c_i = |\mathcal{C}_i|
    \]
    \If{$c_i = 0$}
        \State Mark problem $i$ as unsolved and set $\text{SCPT}_i[k] = 16384$
        \State \textbf{continue}
    \EndIf
    \State Let $\mathcal{T} \gets \text{sorted unique values of } \mathcal{C}_i$
    \ForAll{$k \in \mathcal{K}$}
        \State $s \gets 0$
        \ForAll{$t \in \mathcal{T}$}
            \State $m \gets |\{L \in \mathcal{C}_i \mid L < t\}|$
            \If{$n - m \ge k$}
                \State $p \gets \binom{n-m}{k} / \binom{n}{k}$
                \State $s \gets s + p$
            \EndIf
        \EndFor
        \State $\text{SCPT}_i[k] \gets \min(\mathcal{C}_i) + s$
        \Comment{Using survival form $E[X] = \min + \sum_t \Pr(X > t)$}
    \EndFor
\EndFor

\ForAll{$k \in \mathcal{K}$}
    \State $\text{SCPT}[k] \gets \text{average of } \text{SCPT}_i[k] \text{ over solved problems } i$
\EndFor

\end{algorithmic}
\end{algorithm}

}

\section{Details for experiments}
\subsection{Training Parameters}
\label{Appendix:B.1} 

We train the model on $8 \times$ A800 80GB GPUs with a global batch size of $512$ and a maximum sequence length of $16{,}384$ tokens. To address memory constraints under long-context training, we employ Fully Sharded Data Parallel (FSDP) with CPU offloading.

During optimization, we adopt the GRPO objective with $\beta = 0.001$ and a clipping parameter $\epsilon = 0.2$. In addition, we set the asymmetric upper clipping coefficient $\text{clip\_ratio\_high} = 0.28$ to control excessive policy updates on high-advantage samples. The learning rate is fixed at $1 \times 10^{-6}$. For response generation, we use standard sampling parameters: temperature $= 1.0$ and top-$p = 1.0$.

Training runs for $300$ update steps. At each step, we sample $16$ responses per prompt to construct response groups for GRPO-based advantage computation. Validation is conducted periodically during training, and checkpoints are saved every $10$ steps for model selection and reproducibility.

Notably, we do not incorporate KL divergence regularization into the reward function; the reward signal is computed independently of any KL penalty term.

\subsection{Datasets}
\label{Appendix:B.2} 

\paragraph{DeepScaleR~\cite{deepscaler2025}.}
DeepScaleR is a curated mathematical reasoning dataset comprising approximately 40K problems, constructed by scaling up the difficulty distribution of existing open-source mathematical corpora. It is built upon a systematic difficulty-aware selection pipeline: problems are first scored by estimated difficulty, and then progressively sampled to emphasize challenging instances that demand multi-step derivations, algebraic manipulation, and advanced problem-solving strategies. DeepScaleR serves as both a training corpus for reinforcement learning--based reasoning enhancement and a reference benchmark for evaluating mathematical problem-solving. Its main strengths lie in its carefully calibrated difficulty gradient, broad coverage of mathematical topics spanning algebra, geometry, number theory, and combinatorics, and its demonstrated effectiveness in pushing the reasoning boundaries of small-scale language models.

\paragraph{AIME24~\cite{aime24}.}
AIME24 consists of the official problems from the 2024 American Invitational Mathematics Examination (AIME) I and II and has become a widely used benchmark for evaluating high-level mathematical reasoning. AIME represents the advanced stage of the AMC contest series; its problems typically require sophisticated combinatorial analysis, geometric constructions, or number-theoretic reasoning, and each admits a unique three-digit integer answer. In public repositories, AIME24 usually appears as a 30-problem set designed to assess models under realistic contest conditions. Its main strengths lie in its high difficulty, authoritative sources, and close alignment with real competition settings.

\paragraph{AMC23~\cite{amc23}.}
AMC23 is a benchmark derived from the 2023 American Mathematics Competition, primarily AMC~10 and AMC~12. The problems are reformatted for model evaluation and usually take the form of multiple-choice or integer-answer questions, covering core high-school contest topics such as algebra, geometry, probability, and number theory. The overall difficulty exceeds routine classroom exercises but remains below that of AIME and Olympiad-level problems. Compared with AIME24, AMC23 typically contains a larger number of instances and is therefore commonly used as a medium-difficulty benchmark to evaluate model robustness and generalization on contest-style reasoning tasks.

\paragraph{MATH-500~\cite{hendrycks2021measuring}.}
MATH-500 is a curated subset of 500 competition-level problems drawn from the MATH dataset introduced by Hendrycks et~al. The original MATH dataset contains approximately 12{,}500 U.S. high-school competition problems, each accompanied by a full step-by-step solution---usually written in \LaTeX---and spanning topics such as algebra, geometry, counting and probability, number theory, and precalculus. Difficulty levels follow the Art of Problem Solving (AoPS) scale from~1 to~5, with the highest tier corresponding to AIME-level questions. MATH-500 inherits the original dataset's emphasis on multi-step reasoning and strict answer verification: models are typically required to produce complete derivations and a canonical final answer, and evaluation is conducted using an exact-match criterion. Owing to its high-quality problems and broad topical coverage, it has become a canonical benchmark in mathematical reasoning research.

\paragraph{Minerva Math~\cite{lewkowycz2022solving}.}
Minerva Math (also referred to as the \emph{minerva\_math benchmark}) is a collection of mathematical and quantitative reasoning problems designed around Google Research's Minerva language model to assess formal mathematical and scientific reasoning capabilities. In the original work, \emph{``Solving Quantitative Reasoning Problems With Language Models,''} large-scale technical pre-training and advanced prompting strategies---such as chain-of-thought prompting and majority voting---enabled Minerva to generate multi-step solutions combining numerical computation and symbolic manipulation without relying on external tools, leading to substantial gains across multiple STEM benchmarks. In publicly released evaluations, such as the knoveleng/Minerva-Math dataset, roughly 272 problems are included, primarily sourced from MIT OpenCourseWare and covering advanced undergraduate- and graduate-level STEM topics such as solid-state chemistry, astronomy, differential equations, and special relativity. These benchmarks typically require models to output step-by-step reasoning and a final numerical result. Compared with pure mathematics competition datasets, Minerva Math emphasizes cross-disciplinary quantitative problems and thus provides a complementary lens on real-world scientific and engineering reasoning, although publicly available descriptions of its sampling strategy and difficulty distribution remain relatively limited.

\paragraph{OlympiadBench~\cite{he2024olympiadbench}.}
OlympiadBench, proposed by He et~al.\ in~2024, is a bilingual multimodal Olympiad-level science benchmark designed to evaluate the reasoning and problem-solving capabilities of large language models (LLMs) and large multimodal models (LMMs) using extremely challenging mathematics and physics competition problems. The dataset aggregates 8{,}476 questions drawn from international and national mathematics Olympiads, physics contests, and the Chinese college entrance examination, spanning both text-only and image-based multimodal formats and accompanied by expert-annotated step-by-step solutions to support fine-grained analysis of reasoning processes. Covering both mathematics and physics, OlympiadBench typically requires models to generate detailed solution chains and final answers. Empirical results indicate that even state-of-the-art systems, such as GPT-4V, achieve relatively low performance---around 18\% average accuracy, with even lower scores on physics---highlighting the benchmark's exceptional difficulty and its stringent demands on logical and scientific reasoning. Overall, OlympiadBench provides a testbed far beyond standard high-school competitions and is well suited for probing the upper limits and failure modes of contemporary reasoning models.

\paragraph{MMLU~\cite{hendrycks2020measuring}.}
MMLU (Massive Multitask Language Understanding) is a comprehensive benchmark spanning 57 diverse academic subjects, ranging from STEM disciplines such as mathematics, physics, and computer science to humanities and social sciences including history, law, and philosophy. Each problem is presented in a four-option multiple-choice format, and the questions are drawn from real-world examination materials at difficulty levels from elementary to advanced professional. MMLU has become one of the most widely adopted benchmarks for evaluating the breadth and depth of knowledge acquired by large language models. Its main strengths lie in its broad subject coverage, standardized evaluation format, and ability to probe both factual recall and applied reasoning across highly diverse domains.

\subsection{Detailed Evaluation Protocol}
\label{Appendix:B.3}

\paragraph{Accuracy (Acc)}
For each benchmark dataset, we measure the accuracy as the proportion of correctly answered problems:
\begin{equation}
    \text{Acc} = \frac{1}{N} \sum_{i=1}^{N} \mathbb{1}(\hat{y}_i = y_i),
\end{equation}
where $N$ is the total number of problems in the dataset, $\hat{y}_i$ denotes the model's predicted answer for the $i$-th problem, $y_i$ is the corresponding ground-truth answer, and $\mathbb{1}(\cdot)$ is the indicator function.

\paragraph{Average Token Length (Token)}
We report the average number of tokens generated by the model across all problems in a dataset:
\begin{equation}
    \text{Token} = \frac{1}{N} \sum_{i=1}^{N} l_i,
\end{equation}
where $l_i$ is the number of tokens in the model's response for the $i$-th problem.

\paragraph{Intelligence Per Token (IPT)}
To holistically assess the trade-off between reasoning accuracy and computational cost, following CoThink~\cite{fan2025price},we define \textbf{Intelligence Per Token (IPT)} as:
\begin{equation}
    \mathrm{IPT} = \frac{\mathrm{Acc}}{\mathrm{Token} \;/\; 1000},
\end{equation}
where $\mathrm{Acc}$ denotes the PASS@1 accuracy (\%) and $\mathrm{Token}$ denotes the average output token length. Intuitively, IPT quantifies the accuracy gained per 1K tokens of generation, providing a unified measure of \emph{intelligence density}. A higher IPT indicates that the model achieves stronger reasoning performance with fewer tokens, reflecting a more favorable accuracy--efficiency trade-off. In the main results (Table~\ref{tab:Main-Results}), we report the IPT value averaged across all benchmark datasets for each method.

\paragraph{Average Change Rate: $\Delta$}
To quantify the variation in performance introduced by a method relative to the original base model, we define the change rate of a metric $M$ on a single dataset as:
\begin{equation}
    \text{CR}(M) = \frac{M_{\text{Method}}}{M_{\text{Vanilla}}} - 1,
\end{equation}
where $M \in \{\text{Acc}, \text{Token}\}$ denotes the evaluation metric, $M_{\text{Method}}$ is the metric value of the method-applied model, and $M_{\text{Vanilla}}$ is the corresponding metric value of the original base model. A positive change rate indicates an increase, while a negative value indicates a decrease.

The \textit{Average Change Rate} aggregates the change rates across all $K$ evaluation datasets:
\begin{equation}
    \Delta(M) = \frac{1}{K} \sum_{j=1}^{K} \text{CR}_j(M),
\end{equation}
where $\text{CR}_j(M)$ denotes the change rate on the $j$-th dataset. In our experiments, $K=5$ corresponding to AIME24, AMC23, MATH-500, OlympiadBench, and Minerva. We report $\Delta\text{ACC}$ and $\Delta\text{Token}$ to summarize the overall changes in accuracy and token length, respectively.

\paragraph{Inference Setup}

During inference, the maximum context size is limited to 16,384 tokens, and the sampling temperature is fixed at 0.6 to balance exploration and stability, as suggested in DeepSeek’s model card. For the AIME dataset, due to its relatively small sample size, we repeat the evaluation 16 times and report the averaged results to reduce variance and improve statistical reliability.

\subsection{Detailed Descriptions of Baseline}
\label{Appendix:B.4}
In this section, we provide a detailed overview of the baseline methods included in our comparison.
All of these methods are grounded in RL and have been published or accepted
at top-tier venues after rigorous peer review, spanning from early 2025 to early 2026.
The most recent among them is JET, which has been accepted at ICLR 2026.
Each method addresses the problem of efficient reasoning from a distinct perspective,
and we summarize their key characteristics below.

\textbf{LC-R1}~\cite{cheng2025optimizing} proposes an explicit length penalty reward that imposes a negative
constraint on the token length of generated reasoning sequences, thereby encouraging overall conciseness.
In addition, it introduces a \emph{compression reward} designed to identify and eliminate ineffective
segments within the reasoning process. Although trained on a different dataset,
the authors release two checkpoints: \texttt{LCR1-1.5B} and \texttt{LCR1-7B}.

\textbf{Laser}~\cite{liu2025learn} introduces a \emph{difficulty-aware length penalty reward}
that dynamically adjusts the length constraint based on the estimated difficulty of the input problem.
Furthermore, it proposes a reward mechanism that explicitly encourages the exploration of
incorrect reasoning trajectories, allowing the model to learn from failures.
Two checkpoints are released: \texttt{Laser-DE-L4096-1.5B} and \texttt{Laser-DE-L4096-7B}---both
trained on the same dataset as ours.

\textbf{AutoThink}~\cite{tu2025learning} proposes a \emph{multi-stage reinforcement learning} framework
with stage-wise reward shaping that progressively trains the model to allocate reasoning effort
based on problem difficulty. Stage~1 employs batch-level reward balancing to maintain the coexistence
of both thinking and no-thinking modes; Stage~2 reinforces task accuracy within each mode;
Stage~3 introduces a length-aware reward to prune unnecessary computation.
Through this progressive training, the model learns to adaptively decide whether to engage
in explicit reasoning without any external classifier.
Trained on the same dataset as ours, it provides two checkpoints:
\texttt{Distill-R1-1.5B-AutoThink-Stage3} and \texttt{Distill-R1-7B-AutoThink-Stage3}.

\textbf{AdaptThink}~\cite{zhang2025adaptthink} leverages reinforcement learning to endow the model with
the ability to \emph{bypass the extended reasoning process} for simple problems entirely.
By learning to recognize when deep thinking is unnecessary, the model can respond directly
without invoking the chain-of-thought process, significantly reducing inference cost on easy queries.
Trained on the same dataset as ours, it provides two checkpoints:
\texttt{AdaptThink-1.5B-delta0.05} and \texttt{AdaptThink-7B-delta0.05}.

\textbf{JET}~\cite{han2025your} trains large reasoning models to recognize when sufficient reasoning
has been performed and to \emph{stop overthinking} proactively. Rather than compressing
or penalizing lengthy reasoning after the fact, JET directly optimizes the model's ability to
terminate its reasoning process at the appropriate point, achieving a more natural balance
between reasoning quality and computational efficiency.
It provides two checkpoints:
\texttt{JET-1.5B} and \texttt{JET-7B}.

\section{Detailed Formulation of the Training Objective}
\label{Appendix:C.1} 

In TRiMS, the policy gradient objective is:




\begin{multline}
\label{Eq:GRPO}
J_{\text{TRiMS}}(\theta) = \mathbb{E}_{(q, a) \sim \mathcal{D},\, \{o_i\}_{i=1}^G \sim \pi_{\theta_{\text{old}}}(\cdot|q)} \Bigg[ \\
\frac{1}{\sum_{i=1}^{G} |o_i|}
\sum_{i=1}^G \sum_{t=1}^{|o_i|}
\min \Bigg(
\frac{\pi_\theta(o_{i,t} \mid q, o_{i,<t})}{\pi_{\theta_{\text{old}}}(o_{i,t} \mid q, o_{i,<t})} \hat{A}_{i,t}, \\
\text{clip}\!\Big(
  \frac{\pi_\theta(o_{i,t} \mid q, o_{i,<t})}{\pi_{\theta_{\text{old}}}(o_{i,t} \mid q, o_{i,<t})},\,
  1 - \varepsilon,\;
  1 + \varepsilon_{\text{high}}
\Big) \hat{A}_{i,t}
\Bigg) \\
- \beta\, D_{\text{KL}}\!\big[\pi_\theta \parallel \pi_{\text{ref}}\big]
\Bigg]
\end{multline}

where $\hat{A}_{i,t} = (R_i - \mu_R^{\text{group}}) / \sigma_R^{\text{batch}}$ is the
batch-level normalized advantage (Sec.~\ref{method:Adv}),
$\varepsilon = 0.2$ and $\varepsilon_{\text{high}} = 0.28$ are the asymmetric
clipping coefficients.

\section{Supplementary Experiments}
\label{Appendix:D}

\subsection{Difficulty-Aware Accuracy Analysis}
\label{Appendix:D.1}

\begin{table}[!h]
\centering
\caption{Accuracy comparison between Vanilla and TRiMS across difficulty levels.}
\label{tab:Appendix-difficulty}
\resizebox{\columnwidth}{!}{
\begin{tabular}{lccccc}
\toprule
\textbf{Method} 
& \textbf{1.0} 
& \textbf{2.0} 
& \textbf{3.0} 
& \textbf{4.0} 
& \textbf{5.0}  \\
\midrule
Vanilla 
& 0.9070 
& 0.9333 
& 0.8952 
& 0.7969 
& 0.6567  \\

TRiMS 
& 0.9535 
& 0.9111 
& 0.9429 
& 0.8125 
& 0.6493  \\
\bottomrule
\end{tabular}
}
\end{table}
Table~\ref{tab:Appendix-difficulty} reports per-level accuracy on MATH-500 for the 1.5B model, complementing the token-length analysis in Sec.~\ref{sec:exp-difficulty}. Despite compressing tokens by 78--84\% across all difficulty levels, TRiMS preserves or slightly improves accuracy at every tier. On easier problems (Level 1 and 3), TRiMS yields notable gains of +4.7\% and +4.8\%, respectively, suggesting that the removal of redundant reasoning steps (e.g., hesitation and backtracking, as identified in Sec.~\ref{sec:lexical}) can even benefit answer correctness. On harder problems (Level 4 and 5), accuracy remains essentially unchanged ($\leq$0.7\% deviation), indicating that TRiMS's compression does not erode the essential reasoning steps required for challenging tasks. These results corroborate the main finding of Sec.~\ref{sec:exp-difficulty}: TRiMS achieves difficulty-proportional compression while faithfully preserving the model's problem-solving capability across the full difficulty spectrum, consistent with the MSL convergence behavior established in Secs.~\ref{sec:MSL-Stategy} and~\ref{sec:MSL-Scale}.

\begin{table*}[!t]
\centering
\caption{Comparison of PASS@1 accuracy, average output length (Token), and Intelligence Per Token (IPT) between the TRiMS model and baseline models across all benchmark datasets.}
\label{tab:Appendix-Qwen3}
\resizebox{\textwidth}{!}{%
\begin{tabular}{lccccccccccccc}
\toprule
\multirow{2}{*}{\textbf{Method}} 
& \multicolumn{2}{c}{\textbf{AMC23}} 
& \multicolumn{2}{c}{\textbf{AIME24}} 
& \multicolumn{2}{c}{\textbf{MATH-500}} 
& \multicolumn{2}{c}{\textbf{Minerva}} 
& \multicolumn{2}{c}{\textbf{Olympiad}} 
& \multicolumn{3}{c}{\textbf{Avg}} \\
\cmidrule(lr){2-3} 
\cmidrule(lr){4-5} 
\cmidrule(lr){6-7} 
\cmidrule(lr){8-9} 
\cmidrule(lr){10-11} 
\cmidrule(lr){12-14}
& \textbf{Acc} 
& \textbf{Token} 
& \textbf{Acc} 
& \textbf{Token} 
& \textbf{Acc} 
& \textbf{Token} 
& \textbf{Acc} 
& \textbf{Token} 
& \textbf{Acc} 
& \textbf{Token} 
& $\Delta$\textbf{Acc (\%)} 
& $\Delta$\textbf{Token (\%)} 
& \textbf{IPT} \\
\midrule

\multicolumn{14}{l}{\textbf{Qwen3-4B}} \\
Vanilla & 77.1 & 9060.0 & 57.9 & 11933.0 & 91.6 & 5021.6 & 41.5 & 6673.1 & 58.2 & 9243.0 & -- & -- & 8.8 \\
TRiMS & 80.7 & 1046.2 &53.5 & 1713.6 & 91.0 & 662.2 & 41.9 & 783.0 & 57.9 & 1081.6 & -0.63 & -88.3 & 70.6 \\

\bottomrule
\end{tabular}%
}
\end{table*}

\subsection{Test-Time Scaling Analysis}
\label{Appendix:D.2}

\begin{figure}
\centering
\includegraphics[width=\linewidth]{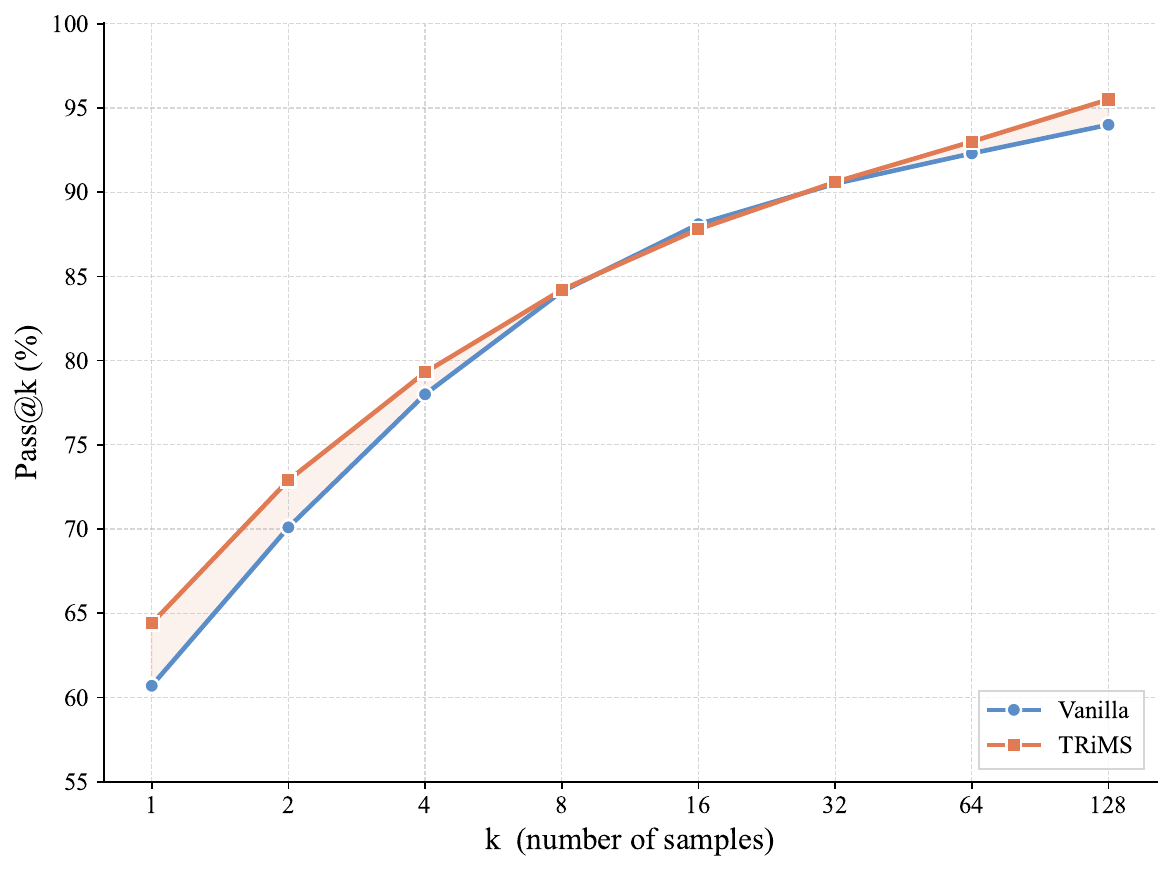}
\caption{PASS@$k$ accuracy on AMC23 (DeepSeek-R1-Distill-Qwen-1.5B) as a function of the number of samples~$k$. The shaded region highlights the accuracy gap between TRiMS and the Vanilla baseline.}
\label{fig:passk}
\end{figure}

Sec.~\ref{sec:efficiency} demonstrates that TRiMS achieves superior accuracy under a fixed budget by fitting more parallel trajectories within the same wall-clock window. Here, we examine the PASS@$k$ metric as a function of independent samples~$k$, decoupled from computational cost.

As shown in Fig.~\ref{fig:passk}, TRiMS consistently achieves higher PASS@$k$ across all budgets, with $k$ ranging from 1 to 128. At $k{=}1$, TRiMS outperforms the Vanilla baseline by approximately 3.7 percentage points (64.5\% vs.\ 60.8\%), indicating each compressed trajectory is individually more likely correct. This advantage persists through moderate $k$ ($k{=}2$--$8$), with the gap remaining visually prominent.

As $k$ increases beyond 32, both models approach full coverage and the gap narrows, as expected since $\text{PASS@}k \to 1$ as $k \to \infty$. Nevertheless, TRiMS maintains a slight edge at $k{=}128$ (95.3\% vs.\ 93.8\%), suggesting its generation distribution assigns higher probability mass to correct solutions.

These results carry two implications. First, TRiMS does not merely compress at the cost of diversity---its per-sample correctness is strictly higher, confirming that compression selectively removes redundant tokens (e.g., hesitation and backtracking, as identified in Sec.~\ref{sec:lexical}) rather than productive reasoning steps. Second, combined with Sec.~\ref{sec:efficiency}, TRiMS delivers a \emph{compounding} efficiency advantage: each sample is both \emph{shorter} (less compute per trajectory) and \emph{more accurate} (fewer samples to reach target coverage). Together, these effects explain the wall-clock savings in Fig.~\ref{fig:parallel_efficiency}, where TRiMS achieves equivalent accuracy with over 93

\subsection{MSL Estimation for Qwen3-4B}
\label{Appendix:D.3}

\begin{figure}[!ht]
    \centering
    \includegraphics[width=\linewidth]{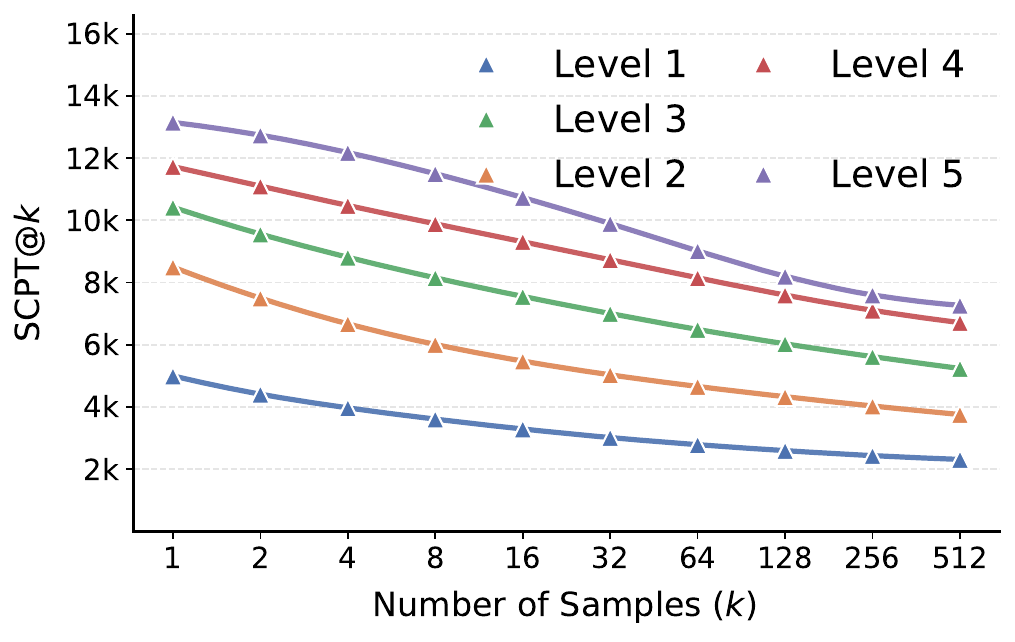}
    \caption{Expected token length of the shortest correct reasoning path for Qwen3-4B across varying difficulty levels as a function of sampling attempts $k$.}
    \label{fig:msl_qwen3}
\end{figure}

To investigate the scaling behavior and generalization of TRiMS on stronger backbones, we empirically estimate the MSL for Qwen3-4B. Following the high-temperature sampling configuration in Appendix~\ref{Appendix:A.1}, we perform offline sampling on the DeepScaleR dataset. For each difficulty level, 30 independent valid trajectories are sampled to analyze the convergence of reasoning lengths.

As illustrated in Fig.~\ref{fig:msl_qwen3}, while Qwen3-4B demonstrates superior reasoning capabilities, it also tends to produce longer and more complex chain-of-thought sequences compared to smaller distillations. Consequently, a static two-stage truncation (2048-4096) is insufficient to capture its intrinsic MSL across all difficulty tiers. To address this, we implement a \text{8192-4096-2048} three-tier adaptive truncation structure:

\begin{equation}
L_{trunc}(Y) = 
\begin{cases} 
2048, & \begin{aligned}[t]
         &\text{if } \exists y_i \in Y,\ |y_i| \leq 2048 \\
         &\text{and } \mathbb{1}_{correct}(y_i)=1
       \end{aligned} \\[6pt]
4096, & \begin{aligned}[t]
         &\text{elif } \exists y_i \in Y,\ |y_i| \leq 4096 \\
         &\text{and } \mathbb{1}_{correct}(y_i)=1
       \end{aligned} \\[6pt]
8192, & \text{otherwise}
\end{cases}
\end{equation}

By expanding the upper truncation boundary to 8192 tokens, we ensure that the model retains sufficient space for complex derivations while still aggressively penalizing redundancy. All other TRiMS components, including Dynamic Batch Aggregation and Batch-level Advantage Normalization, remain identical to the standard implementation described in Sec.~\ref{sec:trims}.

\subsection{Detailed Analysis for Qwen3-4B}
\label{Appendix:D.4}

Based on the empirical results in Table~\ref{tab:Appendix-Qwen3} and the customized three-tier truncation strategy (8192-4096-2048) detailed in Appendix~\ref{Appendix:D.3}, the application of TRiMS to the Qwen3-4B backbone demonstrates robust generalization:

\begin{itemize}
    \item {Significant Token Compression:} TRiMS achieves an average token reduction of 88.3\%, dropping average output length from 9243.0 to 1081.6 tokens.
    \item {Accuracy Preservation:} This aggressive compression incurs a negligible average accuracy penalty of merely -0.63\%.
    \item {Efficiency Surge:} The Intelligence Per Token improves drastically from 8.8 to 70.6.
\end{itemize}

These results confirm that expanding the upper truncation boundary to 8192 tokens successfully allows TRiMS to penalize redundancy while retaining sufficient space for the complex derivations inherent to stronger base models.

\section{Case Study}
\label{Appendix:E}

\begin{figure*}[t]
\centering
\begin{tcolorbox}[colback=white, colframe=black, boxrule=0.8pt, left=0pt, right=0pt, top=0pt, bottom=0pt, arc=0pt, boxsep=0pt]

\begin{tcolorbox}[colback=questionbg, colframe=questionbg, boxrule=0pt, left=5pt, right=5pt, top=5pt, bottom=5pt, sharp corners, before skip=0pt, after skip=0pt]
\textbf{$\blacktriangleright$ Problem:} Find the largest possible real part of \[(75+117i)z+\frac{96+144i}{z}\] where $z$ is a complex number with $|z|=4$.
\end{tcolorbox}

\noindent\rule{\textwidth}{0.4pt}

\begin{tcolorbox}[colback=graybg, colframe=graybg, boxrule=0pt, left=5pt, right=5pt, top=5pt, bottom=5pt, sharp corners, before skip=0pt, after skip=0pt]

\noindent\textcolor{blue!70}{$\bullet$} \textbf{Vanilla:}

\vspace{0.3em}

\texttt{<think>}Alright, so I've got this problem here: I need to find the largest possible real part of the expression $(75 + 117i)z + (96 + 144i)/z$, where $z$ is a complex number with $|z| = 4$. Hmm, okay. Let me try to break this down step by step.
First, since $z$ is a complex number with $|z| = 4$, that means $z$ lies somewhere on the circle in the complex plane with radius 4 centered at the origin. So, I can represent $z$ in polar form, right? Let me recall that any complex number $z$ can be written as $z = re^{i\theta}$, where $r$ is the modulus and $\theta$ is the argument. In this case, $r = 4$, so $z = 4e^{i\theta}$.
Alright, so substituting this into the expression, let's write $z$ as $4e^{i\theta}$. Then, $(75 + 117i)z$ would become $(75 + 117i) \cdot 4e^{i\theta}$, and $(96 + 144i)/z$ would be $(96 + 144i)/(4e^{i\theta})$.
Let me compute each part separately.
\textbf{[...](omit 3450 tokens)}
Therefore, I can be confident that 540 is the correct maximum real part.
Hence, the answer is 540.
\textbf{Final Answer}
$\boxed{540}$
\texttt{</think>}
Given the problem of finding the largest possible real part of the expression $(75 + 117i)z + \frac{96 + 144i}{z}$ where $z$ is a complex number with $|z| = 4$, we start by expressing $z$ in polar form as $z = 4e^{i\theta}$.
\textbf{[...](omit 476 tokens)}
Thus, the largest possible real part of the given expression is: $\boxed{540}$ \noindent\texttt{(4452 tokens)}
\end{tcolorbox}

\noindent\rule{\textwidth}{0.4pt}

\begin{tcolorbox}[colback=greenbg, colframe=greenbg, boxrule=0pt, left=5pt, right=5pt, top=5pt, bottom=5pt, sharp corners, before skip=0pt, after skip=0pt]
\noindent Ground Truth: 540.
\end{tcolorbox}

\noindent\rule{\textwidth}{0.4pt}

\begin{tcolorbox}[colback=graybg, colframe=graybg, boxrule=0pt, left=5pt, right=5pt, top=5pt, bottom=5pt, sharp corners, before skip=0pt, after skip=0pt]
\noindent\textcolor{green!60!black}{$\bullet$} \textbf{TRiMS:}

\vspace{0.3em}

\texttt{<think>}Okay, so I have this problem: I need to find the largest possible real part of (75 + 117i)z + (96 + 144i)/z, where z is a complex number with |z| = 4. Hmm, that seems a bit involved, but let's try to break it down.
\textbf{[...](omit 658 tokens)}
So, $R = \sqrt{104976 + 186624} = \sqrt{291600} = 540$.
Therefore, the maximum value is 540.
\textbf{Final Answer}
$\boxed{540}$
\texttt{</think>}Given the problem, we need to find the largest possible real part of \[(75+117i)z+\frac{96+144i}{z}\]where \(z\) is a complex number with \(|z| = 4\).
First, we express \(z\) as \(4e^{i\theta}\). Then, we compute the real part of the given expression by breaking it down into parts and using trigonometric identities. The real part simplifies to \(324 \cos \theta - 432 \sin \theta\).
This expression can be written as \(R \cos(\theta + \phi)\) where \(R = 540\). Thus, the maximum value is $\boxed{540}$.\noindent\texttt{(995 tokens)}
\end{tcolorbox}

\noindent\rule{\textwidth}{0.4pt}

\begin{tcolorbox}[colback=greenbg, colframe=greenbg, boxrule=0pt, left=5pt, right=5pt, top=5pt, bottom=5pt, sharp corners, before skip=0pt, after skip=0pt]
\noindent Ground Truth: 540.
\end{tcolorbox}

\end{tcolorbox}
\caption{Case study on an AIME problem. Vanilla uses 4452 tokens with extensive step-by-step reasoning, while TRiMS reaches the same correct answer using only 995 tokens.}
\label{fig:case_aime}
\end{figure*}

\begin{figure*}[t]
\centering
\begin{tcolorbox}[colback=white, colframe=black, boxrule=0.8pt, left=0pt, right=0pt, top=0pt, bottom=0pt, arc=0pt, boxsep=0pt]

\begin{tcolorbox}[colback=questionbg, colframe=questionbg, boxrule=0pt, left=5pt, right=5pt, top=5pt, bottom=5pt, sharp corners, before skip=0pt, after skip=0pt]
\textbf{$\blacktriangleright$ Problem:} $\frac{m}{n}$ is the Irreducible fraction value of \[3+\frac{1}{3+\frac{1}{3+\frac13}}\], what is the value of $m+n$?
\end{tcolorbox}

\noindent\rule{\textwidth}{0.4pt}

\begin{tcolorbox}[colback=graybg, colframe=graybg, boxrule=0pt, left=5pt, right=5pt, top=5pt, bottom=5pt, sharp corners, before skip=0pt, after skip=0pt]

\noindent\textcolor{blue!70}{$\bullet$} \textbf{Vanilla:}

\vspace{0.3em}

\texttt{<think>}Okay, so I need to find the value of \( \frac{m}{n} \) where \( \frac{m}{n} \) is the irreducible fraction form of this continued fraction:\(3+\frac{1}{3+\frac{1}{3+\frac{1}{3}}}\). And then find \( m + n \). Alright, let's try to tackle this step by step.

First, I remember that a continued fraction like this can be evaluated by starting from the innermost part and working our way out. So, let me write down the expression again to visualize it better:
\textbf{[...](omit 2400 tokens)}
But, does 109/33 reduce further? Since 109 is prime, and 33 is 3×11, so no, it's irreducible.

Therefore, the answer is m + n = 109 + 33 = 142.

I think that's solid. I don't see any mistakes in my calculations. So, yeah, 142 is the answer.

\textbf{Final Answer}
The value of \( m + n \) is \boxed{142}.
\texttt{</think>}

To find the value of \( \frac{m}{n} \) where \( \frac{m}{n} \) is the irreducible fraction form of the continued fraction:\(3+\frac{1}{3+\frac{1}{3+\frac{1}{3}}}\). we start by simplifying from the innermost part.
\textbf{[...](omit 95 tokens)}
The fraction \( \frac{109}{33} \) is irreducible because 109 is a prime number and does not share any common factors with 33 other than 1. Therefore, \( m = 109 \) and \( n = 33 \).

Thus, the value of \( m + n \) is:\( 109 + 33 = 142 \)
The final answer is \(\boxed{142}\). \noindent\texttt{(3083 tokens)}
\end{tcolorbox}

\noindent\rule{\textwidth}{0.4pt}

\begin{tcolorbox}[colback=greenbg, colframe=greenbg, boxrule=0pt, left=5pt, right=5pt, top=5pt, bottom=5pt, sharp corners, before skip=0pt, after skip=0pt]
\noindent Ground Truth: 142
\end{tcolorbox}

\noindent\rule{\textwidth}{0.4pt}

\begin{tcolorbox}[colback=graybg, colframe=graybg, boxrule=0pt, left=5pt, right=5pt, top=5pt, bottom=5pt, sharp corners, before skip=0pt, after skip=0pt]
\noindent\textcolor{green!60!black}{$\bullet$} \textbf{TRiMS:}

\vspace{0.3em}

\texttt{<think>}Okay, so I need to find the value of this continued fraction: 3 + 1/(3 + 1/(3 + 1/3)). Then express it as an irreducible fraction m/n and find m + n.

Hmm, let's break it down step by step.

Starting from the innermost part: 3 + 1/3.
\textbf{[...](omit 160 tokens)}
Convert 3 to 99/33, so 99/33 + 10/33 = 109/33.

So the continued fraction simplifies to 109/33.

Since 109 is a prime number, I think, and 33 is 3*11, so they have no common factors. So m = 109, n = 33.

Therefore, m + n = 109 + 33 = 142.

\textbf{Final Answer}
\boxed{142}
\texttt{</think>}

To find the value of the continued fraction \(3+\frac{1}{3+\frac{1}{3+\frac{1}{3}}}\) and express it as an irreducible fraction \(\frac{m}{n}\), we start from the innermost part and work our way out.

1. The innermost part is \(3 + \frac{1}{3} = \frac{10}{3}\).
2. Next, we have \(3 + \frac{1}{\frac{10}{3}} = 3 + \frac{3}{10} = \frac{33}{10}\).
3. Finally, we have \(3 + \frac{1}{\frac{33}{10}} = 3 + \frac{10}{33} = \frac{109}{33}\).

The fraction \(\frac{109}{33}\) is irreducible, so \(m = 109\) and \(n = 33\). Therefore, \(m + n = 109 + 33 = 142\).

\(\boxed{142}\) \noindent\texttt{(592 tokens)}
\end{tcolorbox}

\noindent\rule{\textwidth}{0.4pt}

\begin{tcolorbox}[colback=greenbg, colframe=greenbg, boxrule=0pt, left=5pt, right=5pt, top=5pt, bottom=5pt, sharp corners, before skip=0pt, after skip=0pt]
\noindent Ground Truth: 142
\end{tcolorbox}

\end{tcolorbox}
\caption{Case study on a continued fraction problem. Vanilla uses 3083 tokens with repeated verification, while TRiMS solves it in only 592 tokens.}
\label{fig:case_fraction}
\end{figure*}

\begin{figure*}[t]
\centering
\begin{tcolorbox}[colback=white, colframe=black, boxrule=0.8pt, left=0pt, right=0pt, top=0pt, bottom=0pt, arc=0pt, boxsep=0pt]

\begin{tcolorbox}[colback=questionbg, colframe=questionbg, boxrule=0pt, left=5pt, right=5pt, top=5pt, bottom=5pt, sharp corners, before skip=0pt, after skip=0pt]
\textbf{$\blacktriangleright$ Problem:} Convert the point $(0,3)$ in rectangular coordinates to polar coordinates.  Enter your answer in the form $(r,\theta),$ where $r > 0$ and $0 \le \theta < 2 \pi.$
\end{tcolorbox}

\noindent\rule{\textwidth}{0.4pt}

\begin{tcolorbox}[colback=graybg, colframe=graybg, boxrule=0pt, left=5pt, right=5pt, top=5pt, bottom=5pt, sharp corners, before skip=0pt, after skip=0pt]

\noindent\textcolor{blue!70}{$\bullet$} \textbf{Vanilla:}

\vspace{0.3em}

\texttt{<think>}Okay, so I have this problem where I need to convert the rectangular coordinates (0, 3) to polar coordinates. Hmm, I remember that polar coordinates are represented as (r, $\theta$), where r is the distance from the origin and $\theta$ is the angle from the positive x-axis. But let me think through the exact steps so I don't make any mistakes.

First, I recall that the conversion from rectangular (x, y) to polar (r, $\theta$) involves using some trigonometry. Specifically, I think r is calculated using the formula $r = \sqrt{x^2 + y^2}$. Yeah, that seems right because it's essentially the hypotenuse of a right triangle formed by x and y as the legs.
\textbf{[...](omit 1080 tokens)}
Therefore, I feel confident that $(3, \pi/2)$ is the correct polar coordinate for the rectangular point (0, 3).

\textbf{Final Answer}
The polar coordinates are $\boxed{(3, \frac{\pi}{2})}$.
\texttt{</think>}
To convert the rectangular coordinates $(0, 3)$ to polar coordinates, we use the formulas for conversion from rectangular to polar coordinates. 

First, we calculate the radius $r$ using the formula:
\textbf{[...](omit 65 tokens)}
Next, we determine the angle $\theta$. Since the point $(0, 3)$ lies on the positive y-axis, the angle $\theta$ is $\frac{\pi}{2}$ radians.

Thus, the polar coordinates are:
$\boxed{(3, \frac{\pi}{2})}$ \noindent\texttt{(1555 tokens)}
\end{tcolorbox}

\noindent\rule{\textwidth}{0.4pt}

\begin{tcolorbox}[colback=greenbg, colframe=greenbg, boxrule=0pt, left=5pt, right=5pt, top=5pt, bottom=5pt, sharp corners, before skip=0pt, after skip=0pt]
\noindent Ground Truth: $\left(3, \dfrac{\pi}{2}\right)$
\end{tcolorbox}

\noindent\rule{\textwidth}{0.4pt}

\begin{tcolorbox}[colback=graybg, colframe=graybg, boxrule=0pt, left=5pt, right=5pt, top=5pt, bottom=5pt, sharp corners, before skip=0pt, after skip=0pt]
\noindent\textcolor{green!60!black}{$\bullet$} \textbf{TRiMS:}

\vspace{0.3em}

\texttt{<think>}Okay, so I need to convert the rectangular coordinates (0, 3) to polar coordinates. Hmm, I remember that polar coordinates are in the form (r, $\theta$), where r is the distance from the origin and $\theta$ is the angle from the positive x-axis.

First, let's recall the formulas for converting from rectangular (x, y) to polar (r, $\theta$). I think r is calculated as the square root of ($x^2 + y^2$), right?
\textbf{[...](omit 135 tokens)}
Wait, let me double-check. The formulas for $\theta$ are usually arctangent(y/x), but when x is 0, we have to be careful. Since x is 0 and y is positive, $\theta$ is $\pi/2$. So yes, $\theta = \pi/2$.

Therefore, the polar coordinates are $(3, \pi/2)$.

\textbf{Final Answer}
The polar coordinates are $\boxed{\left(3, \dfrac{\pi}{2}\right)}$.
\texttt{</think>}To convert the rectangular coordinates $(0, 3)$ to polar coordinates, we use the formulas for $r$ and $\theta$.

First, calculate $r$:
\[
r = \sqrt{0^2 + 3^2} = \sqrt{9} = 3
\]

Next, calculate $\theta$. Since the point is on the positive y-axis, $\theta = \frac{\pi}{2}$.

Thus, the polar coordinates are $\boxed{\left(3, \dfrac{\pi}{2}\right)}$.\noindent\texttt{(413 tokens)}
\end{tcolorbox}

\noindent\rule{\textwidth}{0.4pt}

\begin{tcolorbox}[colback=greenbg, colframe=greenbg, boxrule=0pt, left=5pt, right=5pt, top=5pt, bottom=5pt, sharp corners, before skip=0pt, after skip=0pt]
\noindent Ground Truth: $\left(3, \dfrac{\pi}{2}\right)$
\end{tcolorbox}

\end{tcolorbox}
\caption{Case study on a coordinate conversion problem. Vanilla uses 1555 tokens with extensive self-verification, while TRiMS solves it in only 413 tokens.}
\label{fig:case_polar}
\end{figure*}

\begin{figure*}[t]
\centering
\begin{tcolorbox}[colback=white, colframe=black, boxrule=0.8pt, left=0pt, right=0pt, top=0pt, bottom=0pt, arc=0pt, boxsep=0pt]

\begin{tcolorbox}[colback=questionbg, colframe=questionbg, boxrule=0pt, left=5pt, right=5pt, top=5pt, bottom=5pt, sharp corners, before skip=0pt, after skip=0pt]
\textbf{$\blacktriangleright$ Problem:} Isosceles trapezoid $ABCD$ has parallel sides $\overline{AD}$ and $\overline{BC},$ with $BC < AD$ and $AB = CD.$ There is a point $P$ in the plane such that $PA=1, PB=2, PC=3,$ and $PD=4.$ Let $\frac{r}{s}=\frac{BC}{AD}$ is irreducible fraction, what is the value of $r+s$?
\end{tcolorbox}

\noindent\rule{\textwidth}{0.4pt}

\begin{tcolorbox}[colback=graybg, colframe=graybg, boxrule=0pt, left=5pt, right=5pt, top=5pt, bottom=5pt, sharp corners, before skip=0pt, after skip=0pt]

\noindent\textcolor{blue!70}{$\bullet$} \textbf{Vanilla:}

\vspace{0.3em}

\texttt{<think>}Okay, so I've got this problem about an isosceles trapezoid \(ABCD\) with parallel sides \(AD\) and \(BC\), where \(BC < AD\). The legs \(AB\) and \(CD\) are equal in length. There's a point \(P\) somewhere in the plane such that the distances from \(P\) to the vertices are \(PA = 1\), \(PB = 2\), \(PC = 3\), and \(PD = 4\). I need to find the ratio \(\frac{BC}{AD}\) expressed as an irreducible fraction \(\frac{r}{s}\) and then compute \(r + s\).

Hmm, okay. Let's start by drawing a diagram to visualize this. I imagine an isosceles trapezoid with the longer base \(AD\) at the bottom and the shorter base \(BC\) at the top. Point \(P\) is somewhere in the plane, and it's connected to all four vertices with those specific distances. 
\textbf{[...](omit 15800 tokens)}
Alternatively, \(a= 16\):

\(16^3=4096\)

\(-25*256= -6400\)

\(64*16=1024\)

\(+900\)

Total:4096 -6400 +1024 +900= (4096 +1024 +900) -6400=6020 -6400= -380$\neq$0

Alternatively, \(a= 17\):

\(17^3=4913\)

\(-25*289= -7225\)

\(64*17=1088\)

\(+900\)

Total:4913 -7225 +1088 +900= (4913 +1088 +900) -7225=6901 -
\textbf{[Generation Error: Limit Reached]} \noindent\texttt{(16384 tokens)}
\end{tcolorbox}

\noindent\rule{\textwidth}{0.4pt}

\begin{tcolorbox}[colback=greenbg, colframe=greenbg, boxrule=0pt, left=5pt, right=5pt, top=5pt, bottom=5pt, sharp corners, before skip=0pt, after skip=0pt]
\noindent Ground Truth: 4
\end{tcolorbox}

\noindent\rule{\textwidth}{0.4pt}

\begin{tcolorbox}[colback=graybg, colframe=graybg, boxrule=0pt, left=5pt, right=5pt, top=5pt, bottom=5pt, sharp corners, before skip=0pt, after skip=0pt]
\noindent\textcolor{green!60!black}{$\bullet$} \textbf{TRiMS:}

\vspace{0.3em}

\texttt{<think>}Okay, so I have this problem about an isosceles trapezoid ABCD with bases AD and BC, where BC is shorter than AD. AB equals CD, so it's an isosceles trapezoid. There's a point P somewhere such that PA=1, PB=2, PC=3, PD=4. I need to find the ratio BC/AD in simplest terms and then compute r+s where r/s is that ratio.

Hmm, trapezoid with point P inside or outside with given distances. I remember something about using coordinates or maybe the British flag theorem, but since it's a trapezoid, maybe coordinate geometry can help.
\textbf{[...](omit 2800 tokens)}
Alternatively, maybe the ratio is 1/2.

But not sure. Maybe the answer is 1/3, so r+s=4.

Wait, maybe the answer is 1/3, so 1+3=4.

\textbf{Final Answer}
\boxed{4}
\texttt{</think>}

Given an isosceles trapezoid \(ABCD\) with parallel sides \(\overline{AD}\) and \(\overline{BC}\), where \(BC < AD\) and \(AB = CD\). There is a point \(P\) such that \(PA = 1\), \(PB = 2\), \(PC = 3\), and \(PD = 4\). We need to find the ratio \(\frac{BC}{AD}\) and compute \(r + s\) where the ratio is irreducible.

Using coordinate geometry and solving the system of equations, we find the ratio \(\frac{BC}{AD}\) is \(\frac{1}{3}\). Thus, \(r = 1\) and \(s = 3\), giving \(r + s = 4\).\(\boxed{4}\) \noindent\texttt{(3121 tokens)}
\end{tcolorbox}

\noindent\rule{\textwidth}{0.4pt}

\begin{tcolorbox}[colback=greenbg, colframe=greenbg, boxrule=0pt, left=5pt, right=5pt, top=5pt, bottom=5pt, sharp corners, before skip=0pt, after skip=0pt]
\noindent Ground Truth: 4
\end{tcolorbox}

\end{tcolorbox}
\caption{Case study on a geometry problem. Vanilla exhausts the 16384 token limit without reaching an answer, while TRiMS correctly solves it in 3121 tokens.}
\label{fig:case_trapezoid}
\end{figure*}

\end{document}